\pdfoutput=1
\documentclass[11pt]{article}
\usepackage{tabularx}
\usepackage{booktabs}

\usepackage[final]{acl}
\usepackage{rotating}
\usepackage{booktabs}
\usepackage{graphicx}
\usepackage{lscape}
\usepackage{times}
\usepackage{latexsym}
\usepackage[T1]{fontenc}
\usepackage[utf8]{inputenc}
\usepackage{multirow}

\usepackage{microtype}
\usepackage{array}
\usepackage{inconsolata}

\usepackage{graphicx}
\usepackage{booktabs}

\title{\textsc{PerCul}: A Story-Driven Cultural Evaluation of LLMs in Persian}

\author{
 \textbf{Erfan Moosavi Monazzah*\textsuperscript{2,3}},
 \textbf{Vahid Rahimzadeh*\textsuperscript{1,3}}, \\
 \textbf{Yadollah Yaghoobzadeh\textsuperscript{1,3}}, 
 \textbf{Azadeh Shakery\textsuperscript{1,4}}, and
 \textbf{Mohammad Taher Pilehvar\textsuperscript{5}}
\\
 \textsuperscript{1}University of Tehran, Iran \\
 \textsuperscript{2}Iran University of Science and Technology, Tehran, Iran\\
 \textsuperscript{3}Tehran Institute for Advanced Studies, Khatam University, Iran \\
  \textsuperscript{4}Institute for Research in Fundamental Sciences (IPM), Tehran, Iran \\
  \textsuperscript{5}Cardiff University, United Kingdom 
\\
 \texttt{moosavi\_m@comp.iust.ac.ir, rahimzade@ut.ac.ir}
}

\begin{document}
\maketitle

\begin{abstract}
Large language models predominantly reflect Western cultures, largely due to the dominance of English-centric training data. 
This imbalance presents a significant challenge, as LLMs are increasingly used across diverse contexts without adequate evaluation of their cultural competence in non-English languages, including Persian. 
To address this gap, we introduce \textsc{PerCul}, a carefully constructed dataset designed to assess the sensitivity of LLMs toward Persian culture. 
\textsc{PerCul} features story-based, multiple-choice questions that capture culturally nuanced scenarios.
Unlike existing benchmarks, \textsc{PerCul} is curated with input from native Persian annotators to ensure authenticity and to prevent the use of translation as a shortcut. 
We evaluate several state-of-the-art multilingual and Persian-specific LLMs, establishing a foundation for future research in cross-cultural NLP evaluation. Our experiments demonstrate a 11.3\% gap between best closed source model and layperson baseline while the gap increases to 21.3\% by using the best open-weight model. You can access the dataset from here:
\href{https://huggingface.co/datasets/teias-ai/percul}{https://huggingface.co/datasets/teias-ai/percul}
\end{abstract}

\section{Introduction}

\begingroup
  \hypersetup{hidelinks}       % Disable hyperlinks
  \renewcommand{\thefootnote}{*} % Use * symbol
  \footnotetext{Equal contribution, ordered randomly}
\endgroup

Effective interactions between users from diverse backgrounds and LLMs are contingent on outputs that are culturally relevant~\cite{intro_2024_0}. As the use of generative artificial intelligence increases to expedite and automate personal and professional tasks, the cultural values embedded in AI models may inadvertently bias people's authentic expression and perpetuate the dominance of certain cultures~\cite{intro_2024_3}, particularly Western culture, which is over-represented in English-dominated training data~\cite{intro_2024_1, beer_2024}. This highlights the importance of creating culture-specific benchmarking tools to assess the extent to which LLMs encapsulate knowledge about particular cultures.

\begin{figure}[t]
  \includegraphics[width=\columnwidth]{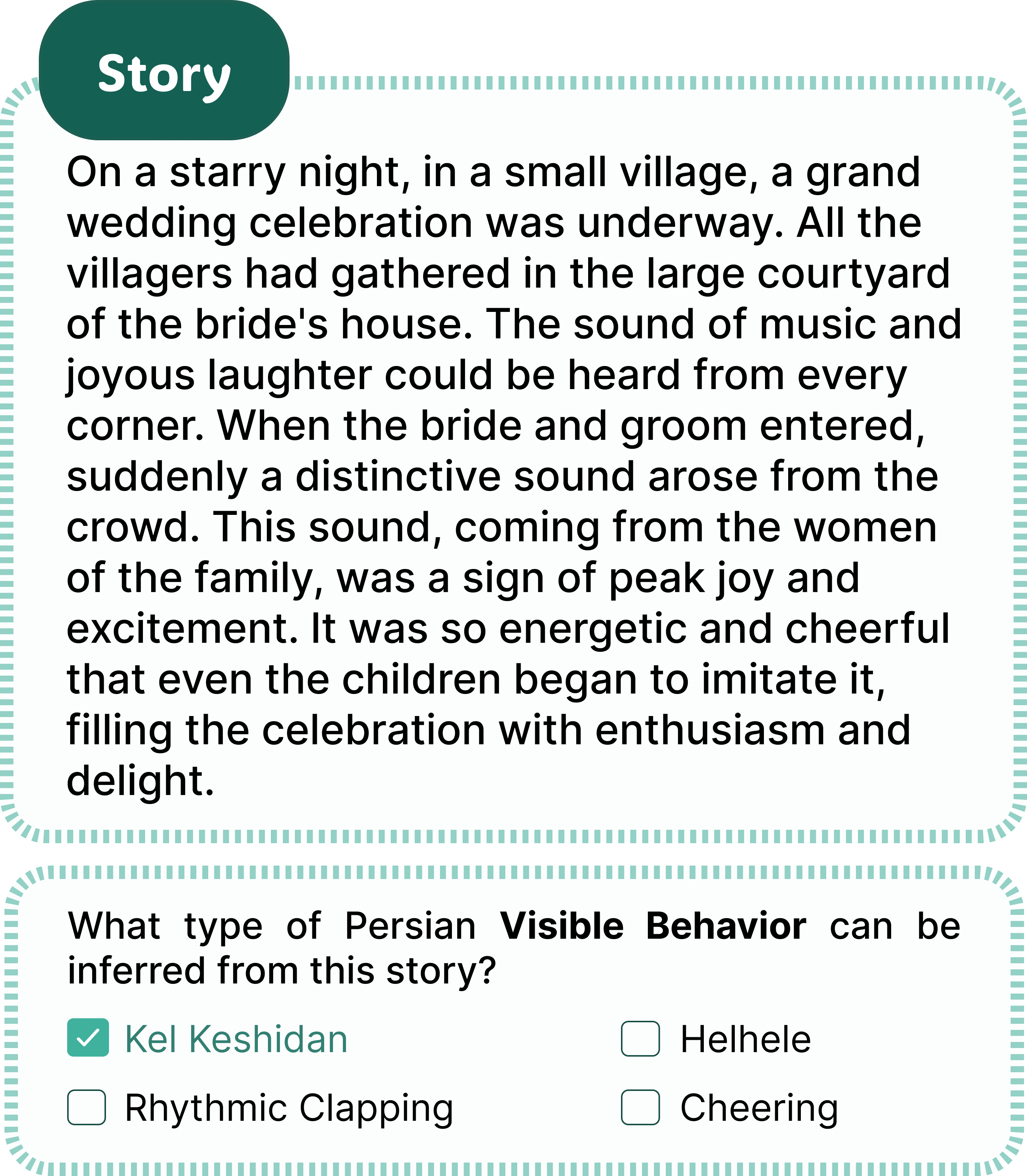}
  \caption{A translated example of \textsc{PerCul}, implying a cultural concept in Visible Behavior category.}
  \label{fig:example}
\end{figure}
Despite the numerous benchmarks that evaluate various aspects of LLMs~\cite{llm_evaluation}, a significant gap remains in assessing their knowledge of culture across many non-English languages such as Persian. Although some efforts have been made to create LLM benchmarks for the Persian language, focusing on reading comprehension and scientific knowledge~\cite{khayyam_2024,pquad_2023, parsinlu_2021}, cultural benchmarks specifically tailored for Persian are limited, either concentrating on specific aspects such as social norms~\cite{psn_2024, blend_2024} or being constrained by size~\cite{cultural_en_5_2024}.

To address the gap in evaluating the sensitivity of LLMs to Persian culture, we introduce \textsc{PerCul}, a carefully curated dataset featuring multiple-choice questions.
In \textsc{PerCul}, cultural concepts are subtly embedded in short stories (see Figure~\ref{fig:example}). Cultural phenomena often manifest implicitly through interactions between individuals, which can be effectively conveyed through the medium of a short story~\cite{story_2021,culture_in_dialogue}. 
Unlike previous benchmarks, \textsc{PerCul} is specifically curated for Persian demographic, avoiding irrelevant or overly generalized concepts shared by other cultures~\cite{psn_2024,blend_2024}. 
Furthermore, the dataset is resistant to use translation as a proxy~\cite{trans_2024,trans_2021}.

Unlike other similar datasets that rely on LLMs for generation~\cite{acegpt_2024,psn_2024}, \textsc{PerCul} leverages input from diverse native Persian annotators, ensuring broader knowledge coverage.
LLMs are used only for generating storylines based on our handcrafted data, with human editing involved to ensure factual accuracy and to prevent hallucinations.

To establish baselines, we evaluate several recent models from different families, including Meta Llama 3.x~\cite{llama3}, OpenAI GPT~\cite{model_gpt4}, Anthropic Claude~\cite{model_sonnet}, as well as state-of-the-art Persian-specific models, namely PersianMind-v1.0~\cite{persianmind} and Dorna-Llama3-8B-Instruct~\cite{dorna}.
Our experiments reveal a gap between models understanding of Persian culture and layperson baseline.
We also demonstrate that translating \textsc{PerCul} results in a significant drop in model's performance. 
Furthermore, we observe that Persian fine-tuned LLMs perform worse than their respective multilingual base models, which may result from a low-quality, small training set.
Lastly, our comprehensive error analysis highlights a limitation in LLMs: they often rely on surface-level details rather than synthesizing contextual clues when it comes to identifying specific cultural concepts.

\section{Related Work}
LLM evaluation has expanded significantly in recent years, covering aspects such as reasoning~\cite{llm_bench_1_rm, llm_bench_4_rm}, knowledge and language understanding~\cite{llm_bench_3_k, llm_bench_5_k}, and instruction following~\cite{llm_bench_0_if, llm_bench_6_if}. As LLMs have dramatically improved in capability, the focus of benchmarking has shifted towards more challenging tasks, such as cultural awareness. Despite numerous attempts to develop cultural benchmarks for English~\cite{cultural_en_1_2024, cultural_en_2_2024, cultural_en_4_2024, cultural_en_5_2024} and other widely spoken languages~\cite{blend_2024, cultural_us_egypt_2024, cultural_multi_1_2024, cultural_korean_2024, acegpt_2024, cultural_en_3_2023}, a gap remains in evaluations of less-studied languages and cultures, such as Persian.

Most existing Persian benchmarks focus on language understanding tasks such as textual entailment and question answering~\cite{farstail_2023, pquad_2023, parsquad_2021, parsinlu_2021}, or the evaluation of factual/scientific knowledge of LLMs~\cite{khayyam_2024, persian_bench_2024}. 
For instance, the Khayyam-Challenge~\cite{khayyam_2024} proposes a set of 20K Persian questions divided into 38 tasks, but these tasks are mainly school-level examinations, primarily covering mathematical and scientific subjects. Although this is useful for evaluating the capabilities of LLMs to solve scientific problems in Persian, it fails to assess LLMs' understanding of Persian culture. This also applies to the work of~\cite{persian_bench_2024} which introduces two new datasets to evaluate LLM abilities in solving Persian mathematical and scientific questions.

Among works on Persian culture, PSN~\cite{psn_2024} provides pairs of social norms and contexts along with a label for each pair describing the appropriateness of each pair. However, it is limited to social norms, leaving out other important aspects such as \textit{Visible Behavior} or \textit{Rituals}. BLEnD~\cite{blend_2024} and CulturalBench~\cite{cultural_en_5_2024} are multi-cultural datasets that despite the inclusion of certain questions about Persian culture, present crucial limitations. BLEnD primarily features questions that focus on non-Persian cultural events and traditions, such as {\it Thanksgiving} or {\it Christmas}, making it less relevant for assessing cultures where these events are not celebrated, such as Persian. CulturalBench, while contains question relevant to Persian culture, is small in size.

\begin{figure*}[!ht]
  \centering
  \includegraphics[width=\textwidth]{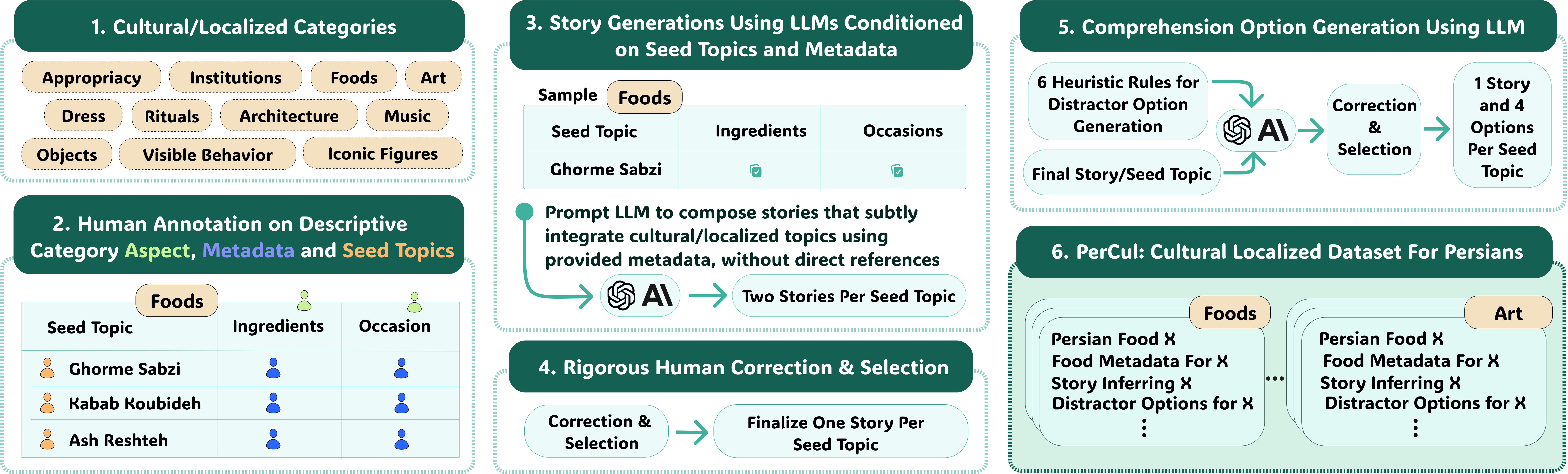}
  \caption{\textsc{PerCul} was generated through a stepwise process: (1) identifying cultural categories using Hall's Triad of Culture, (2) native annotators generating facets, topics, and metadata, (3) generating storylines with capable LLMs, (4) rigorous human correction and selection of stories, (5) creating comprehension options with heuristic rules, and (6) dataset compilation.}
  \label{fig:dataset_generation}
\end{figure*}

\section{\textsc{PerCul} Construction}

The process of creating \textsc{PerCul} consists of multiple steps, as shown in Figure~\ref{fig:dataset_generation}.
Briefly, the creation process begins (1) by identifying cultural categories based on Hall's Triad of Culture~\cite{katan}. (2) Then, native annotators generate descriptive facets, seed topics, and metadata for these categories. (3) Using this metadata, LLMs produce storylines. (4) These storylines undergo rigorous human correction and selection. (5) LLMs also create comprehension options guided by careful human-crafted heuristic rules, followed by human correction and selection. 
The resulting dataset features culturally relevant Persian story comprehension questions in multiple-choice format. These questions subtly incorporate cultural elements from various categories, informed by human-generated metadata, without directly referencing the cultural concepts.

\subsection{Base Theory}
To effectively assess cultural understanding of LLMs, we must first establish a clear definition of culture. 
One widely accepted definition is Edward T. Hall's Triad of Culture, commonly known as \textit{Cultural Iceberg Theory}.
This model, which is frequently used by intercultural scholars and trainers~\cite{katan, halls_ref_2013, halls_ref_2019}, has recently gained traction among NLP researchers~\cite{halls_ref_2024}.
Hall’s theory classifies culture into three levels: technical, formal, and informal.
The technical level is characterized by empirical facts and precise definitions, typical in scientific discourse. 
The formal level consists of traditions and social norms that shape everyday life, often going unnoticed unless violated. 
The informal level, meanwhile, encompasses unconscious, emotionally driven behaviors absorbed through socialization~\cite{katan}. 
\begin{figure}[t]
  \includegraphics[width=\columnwidth]{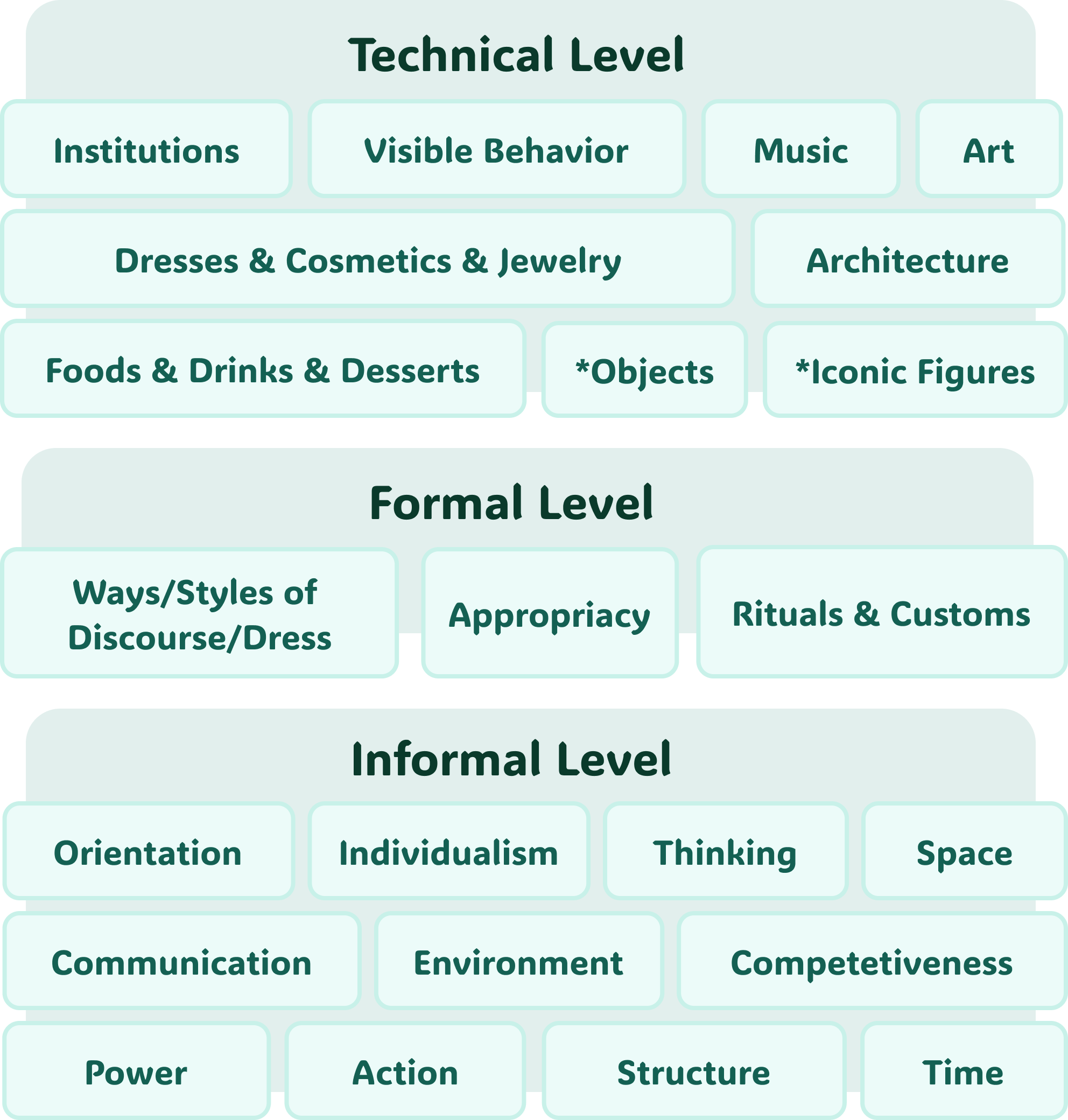}
  \caption{Hall's triad of cultural levels, (*) indicates our extensions to the categories.}
  \label{fig:halls}
\end{figure}
Our analysis centers on the technical and formal level (see Figure~\ref{fig:halls}). We choose not to include the informal category due to the difficulty of capturing these unconscious elements in text and the challenges of collecting data on implicit behaviors from large populations. To adapt Hall's Triad to Persian culture, we expanded the technical level to include Iconic Figures and Objects.

\begin{table*}[t!]
\centering
\small
\setlength{\tabcolsep}{6pt}
\begin{tabularx}{\textwidth}{l c c l X}
\toprule
\textbf{Category} & \textbf{Edition} & \textbf{\# Stories} & \textbf{Sample Topic} & \textbf{Facets} \\
\midrule
Institution & 21.9\% & 43 & Guardian Council & Purpose and Function, Reason of Establishment\newline Organizational Structure, Social Impact, Location \\ 
\midrule
Music & 0.3\% & 32 & Bandari Music & Musical Instruments, Vocal Styles\newline Dominant Themes, Performing Occasions \\ 
\midrule
Dress & 1.7\% & 33 & Charqad & Appropriate occasions, Gender, Materials\newline Colors and Patterns, Regional variations \\ 
\midrule
Objects & 5.3\% & 42 & Aftabeh &  Purpose or Function, Historical Context\newline Materials, Related Customs or Objects \\ 
\midrule
Vis. Behavior & 16.9\% & 56 & Cross Legged Sitting & Environmental elements (time of day, location, etc.)\newline Sample Situations \\ 
\midrule
Art & 10.3\% & 32 & Khatam & Historical background, Dominant color palette\newline Material used, Related ceremonies or customs \\ 
\midrule
Iconic Figures & 15.7\% & 55 & Amir Kabir & Appearance or distinguishing features\newline Cultural and social influences \\ 
\midrule
Appropriacy & 17.0\% & 36 & Couples Kissing in Public & Environment, Context, Social Expectation \\ 
\midrule
Rituals & 4.0\% & 29 & Chaharshanbe Suri & Purpose or importance, Participants and roles\newline Steps/Parts or Tools, Beliefs/Superstitions \\ 
\midrule
Architecture & 14.1\% & 43 & Qanat & Historical period, Symbolism/Significance\newline Materials, Location, Purpose of structure \\ 
\midrule
Foods & 16.9\% & 191 & Ghormeh Sabzi & Ingredients, Preparation Methods, Hot/Cold\newline Food/Drink Pairings, Occasions of Use \\ 
\bottomrule
\end{tabularx}
\caption{An overview of cultural categories: \textit{Edition (\%)} represents the percentage of tokens that were changed during human editing of the stories, \textit{\# Stories} indicates the number of stories in each category, and \textit{Sample Topic} \& \textit{Facets} provide examples of cultural topics and facets used to collect metadata, respectively.}
\label{tab:topics_facets}
\end{table*}

\subsection{Seed Topic Collection}
To represent Persian culture across all categories depicted in Figure~\ref{fig:halls}, we collected 709 cultural seed topics through human annotation (see Table~\ref{tab:topics_facets}).
A group of native Persian speakers from diverse cultural backgrounds (see Appendix~\ref{appendix:annotators}) contributed their perspectives while following the guidelines in Appendix~\ref{appendix:guideline}. 
To ensure the quality of the seed topics, an inter-agreement assessment was conducted among a different group of annotators, and only those seed topics with complete agreement were selected.
This resulted in 556 topics being selected for the final dataset. 
For the appropriacy category, we use sample data from the PSN benchmark~\cite{psn_2024} as seed topics. 
PSN originally contains 1,760 samples, where each sample consists of a social norm, a context for the norm, and a label that describes whether the social norm is \textit{normal}, \textit{taboo}, or \textit{expected} in the provided context.
We carefully selected 36 samples, as numerous entries were similar or differed only in context (despite being context independent). See Table~\ref{tab:topics_facets} for sample seed topics.

\subsection{Metadata Collection} 
To enhance narrative generation and prevent hallucination issues when incorporating LLMs, we collect a set of carefully annotated data from native Persians to ground the generated storylines in next steps in factual infromation.
First, for each category, we collect human-annotated facets (see Table~\ref{tab:topics_facets}). 
These facets were expected to effectively describe the characteristics of the seed topics in that category. They were also required to provide sufficient clues and factual information, allowing inference of the seed topic from an indirectly reflecting narrative. 
Through inter-agreement, we select the final facets that best met these requirements (see Table ~\ref{tab:topics_facets} for a list of facets per category). 
Once finalized, the annotators use the corresponding guidelines in Appendix~\ref{appendix:guideline} to collect category-specific metadata for each seed topic. 
Annotators are encouraged to rely primarily on their personal cultural knowledge, and while Internet search is not prohibited, they are advised to use it only when necessary. This approach helps minimize potential overlap with LLM training data.

\begin{table*}[ht]
\centering
\small
\begin{tabularx}{\textwidth}{X X X}
\toprule
\multicolumn{3}{l}{\textbf{Example 1}} \\
\cmidrule(lr){1-1}
\multicolumn{3}{l}{\textbf{Story:}} \\
\multicolumn{3}{p{\dimexpr\textwidth-2\tabcolsep}}{
    Maryam entered the house with excitement. Her mother greeted her with a smile and said, ``My daughter, you must be tired. How was the exam?'' Maryam happily replied, ``Mom, you won't believe it! I got the first place in the class!'' Her mother hugged her joyfully and said, ``Thank God!'' Then she went to the kitchen and returned with a small container. A pleasant aroma filled the air, and a gentle smoke filled the room. Her mother circled around Maryam, silently reciting a prayer. Maryam felt an unusual sense of calm and gave a heartfelt smile.
} \\
\cmidrule(lr){1-1}
\multicolumn{3}{l}{\textbf{Correct Answer:}} \\
\multicolumn{3}{p{\dimexpr\textwidth-2\tabcolsep}}{
    Mother burned espand.
} \\
\midrule
\textbf{R1: Partial Correctness} & \textbf{R2: Misinterpretation} & \textbf{R3: Unrelated Fact} \\
Mother lit an incense stick. & Mother cooked food. & Mother hugged her. \\
\midrule
\textbf{R4: Plausible Unsupported} & \textbf{R5: Noun Confusion} & \textbf{R6: Overgeneralization} \\
Mother held a prayer ceremony. & Maryam burned espand. & Mother always burns espand. \\
\specialrule{1pt}{2pt}{4pt} 
\multicolumn{3}{l}{\textbf{Example 2}} \\
\cmidrule(lr){1-1}
\multicolumn{3}{l}{\textbf{Story:}} \\
\multicolumn{3}{p{\dimexpr\textwidth-2\tabcolsep}}{
    The grandmother carefully and delicately took the old china teapot out of the cabinet. With a kind smile, she poured the dry tea leaves into the teapot and then reached her hand towards the small container by the samovar. With her fingers, she picked a few small and fragrant seeds and gently dropped them into the teapot. A pleasant aroma filled the space. The grandmother poured the boiling water over the tea and closed the lid of the teapot. After a few minutes, she filled the small cups, and the delightful scent of freshly brewed tea spread throughout the house. The grandchildren eagerly approached the table, eager to drink their grandmother's delightfully aromatic and flavorful tea.
} \\
\cmidrule(lr){1-1}
\multicolumn{3}{l}{\textbf{Correct Answer:}} \\
\multicolumn{3}{p{\dimexpr\textwidth-2\tabcolsep}}{
    Cardamom.
} \\
\midrule
\textbf{R1: Partial Correctness} & \textbf{R2: Misinterpretation} & \textbf{R3: Unrelated Fact} \\
Saffron & Rosewater & Samovar \\
\midrule
\textbf{R4: Plausible Unsupported} & \textbf{R5: Noun Confusion} & \textbf{R6: Overgeneralization} \\
Green Tea & Grandfather & Spices \\
\bottomrule
\end{tabularx}
\caption{Examples of two translated stories with their correct answers and distractor options (R1 to R6).}
\label{tab:example_distractors}
\end{table*}

\subsection{Instance Generation}
Using seed topics and their corresponding metadata, we conduct a semi-automatic process by prompting two state-of-the-art LLMs, GPT 4o ~\cite{model_gpt4} and Claude Sonnet 3.5 ~\cite{model_sonnet}, to generate short storylines leveraging the provided prompts (see Appendix~\ref{appendix:prompts}). 
These storylines imply the respective seed topic using its metadata as clues. 
Table~\ref{tab:example_distractors} shows two samples (translated) from the dataset.
To ensure cultural authenticity and accuracy, two human annotators review and revise the model-generated stories without knowing their source models.
Using the user interface in Appendix~\ref{appendix:uis}, they either rewrite or select the version that best represent the seed topic without direct reference. 
This involves editing, adding, or removing information from the stories, and occasionally, complete rewrites. 
Table~\ref{tab:topics_facets} presents statistics that highlight the extent of the editing carried out in the process.

\subsection{Distractor Options}

We develop six heuristic rules to guide comprehension option generation and use GPT 4o and Sonnet 3.5 to create 24 options per question (2 models $\times$ 6 rules $\times$ 2 options). The options undergo a three-stage selection process:

\begin{enumerate}
    \item \textbf{Initial Selection:} Human annotators evaluate 4 options per heuristic rule (2 from each model) and select the 2 best options that match the rule's intended objective. Model names are hidden to prevent bias.
    
    \item \textbf{Focused Pruning:} From the remaining options, annotators select 6 options per story, allowing up to 2 options from the same rule.
    \\
    \item \textbf{Final Refinement:} Annotators select 3 final options, prioritizing contextual relevance and story alignment.
\end{enumerate}
Each stage includes inter-agreement assessment to validate annotator consistency (see Appendix~\ref{appendix:prompts}). Example of resulting distractors and stories are shown in Table~\ref{tab:example_distractors}.
\begin{table}[t!]
\centering
\small
\setlength{\tabcolsep}{10pt}
\begin{tabular}{llc}
\toprule
& \textbf{Model} & \textbf{Macro Acc.} \\
\midrule
% Claude Models
\multirow{8}{*}{\rotatebox[origin=c]{90}{\bf Closed Source}} &
Claude-3-Haiku   & 0.587 \\
& Claude-3-Sonnet   & 0.680 \\
& Claude-3.5-Sonnet   & \textbf{0.817} \\
& Claude-3-Opus   & 0.793 \\
\cmidrule(lr){2-3}
% GPT Models
& GPT-4o-Mini   & 0.642 \\
& GPT-4o   & \textbf{0.800} \\
\cmidrule(lr){2-3}
% Gemini Models
& Gemini-Flash-1.5   & 0.731 \\
& Gemini-Pro-1.5   & \textbf{0.799} \\
\midrule
% LLaMA Models
\multirow{14}{*}{\rotatebox[origin=c]{90}{\bf Open Weight}} & 
LLaMA-3.2-1B-Inst   & 0.064 \\
& LLaMA-3.2-3B-Inst   & 0.261 \\
& LLaMA-3.1-8B-Inst   & 0.444 \\
& LLaMA-3.1-70B-Inst   & 0.673 \\
& LLaMA-3.1-405B-Inst   & \textbf{0.717} \\
\cmidrule(lr){2-3}
% Gemma Models
& Gemma-2-2B-IT   & 0.348 \\
& Gemma-2-9B-IT   & \textbf{0.675} \\
& Gemma-2-27B-IT   & 0.668 \\
\cmidrule(lr){2-3}
% Cohere Models
& Aya-23-8B   & 0.409 \\
& Command-R-Plus   & \textbf{0.710} \\
\cmidrule(lr){2-3}
% Mistral Models
& Mistral-7B-Instruct-v0.3   & 0.149 \\
& Mistral-Nemo   & \textbf{0.448} \\
& Mixtral-8x22B-Instruct   & 0.388 \\
\cmidrule(lr){2-3}
% Qwen Models
& Qwen-2.5-72B-Instruct   & \textbf{0.619} \\
\midrule
\multicolumn{3}{l}{\bf Persian Fine-Tuned Models} \\
\midrule
& PersianMind v1.0  & 0.033 \\
& Dorna-LLaMA3-8B-Instruct  & \textbf{0.440} \\
\midrule
\multicolumn{2}{l}{\textit{Human Performance}} & 0.930 \\
\bottomrule
\end{tabular}
\caption{The accuracy of different LLMs from different types and families on the dataset. We report macro accuracy across the categories. Models are divided into three types: closed- and open-weight, and Persian-specific.}
\label{tab:model_comparison}
\end{table}
\subsection{Data Statistics}
The final dataset is a comprehensive collection of 592 multiple-choice question-answer pairs, carefully designed to assess story comprehension while subtly incorporating cultural seed topics through short stories without explicit mention. The distribution of these stories across various cultural categories is presented in Table~\ref{tab:topics_facets}, providing a detailed breakdown of the dataset's composition. 
Furthermore, the dataset is accompanied by a set of metadata, which will be made available to ensure a comprehensive understanding of the data and its cultural nuances. 
This metadata will serve as an invaluable resource for researchers and practitioners working with the dataset.
\section{Experiments} 
\begin{figure}[t!]
    \centering
    \includegraphics[width=\columnwidth]{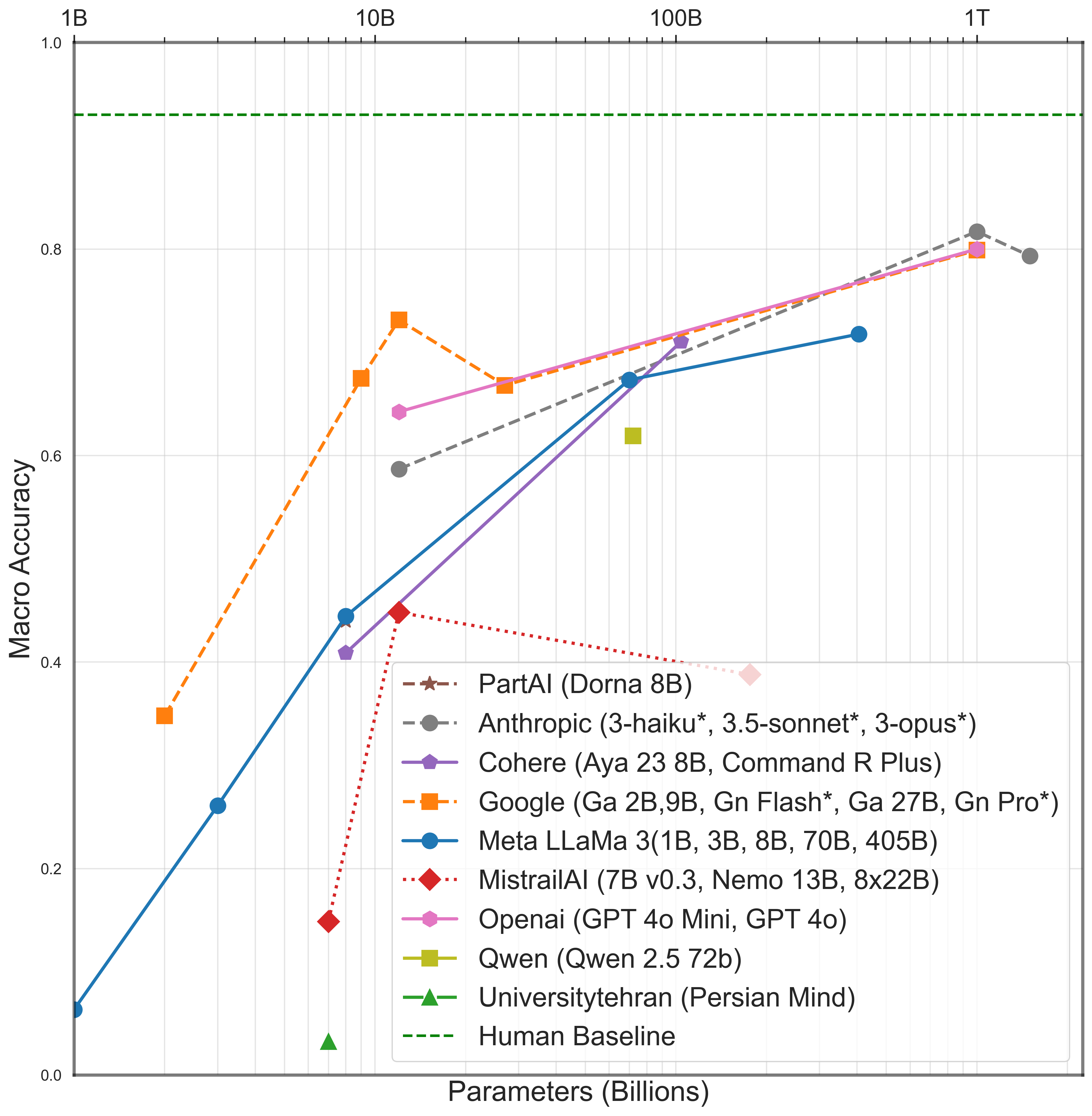}
    \caption{The accuracy on \textsc{PerCul} for different families of models against their number of parameters.}
    \label{fig:acc_experiment}
\end{figure}

We perform a comprehensive series of evaluations on our dataset using state-of-the-art closed-source and open-weight models, as depicted in Figure ~\ref{fig:acc_experiment}. 
We also assess two Persian open-weight LLMs, namely PersianMind-v1.0~\cite{persianmind} and PartAI Dorna-Llama3-8B-Instruct~\cite{dorna}, which are aimed to enhance Persian language and cultural understanding by further pretraining \& fine-tuning on corpora with dominant Persian data. 
To ensure the reproducibility of our experiments, all models utilize zero temperature and allowed to generate up to their maximum generation length. 
We employ the same prompts for each question across all models, which can be found in Appendix~\ref{appendix:prompts}.\footnote{Most of the models are evaluated using their APIs. In instances where a model is not hosted on an API service, it is deployed on a NVIDIA GeForce RTX 3090 GPU and load with either BF16 or FP16 precision.} 
The results are presented in Table~\ref{tab:model_comparison}.\footnote{Given the dataset's balanced nature, macro and micro accuracy metrics are in a similar range.}
According to our findings, the best-performing model on \textsc{PerCul} is Anthropic Sonnet 3.5, with an accuracy of 81.7\% (which is still 11.3\% lower than the human baseline of 93\%).
No open-weight model is present among top five best-performing models. The best performing open-weight model is LlaMA-3.1-405B-Inst with a performance of 71.7\%.
The average accuracy for closed-source and open-weight models are 68.5\% and 40.7\%, respectively.

\begin{figure*}[t!]
    \centering
    \includegraphics[trim={0 1cm 4cm 0},clip,width=0.8\textwidth]{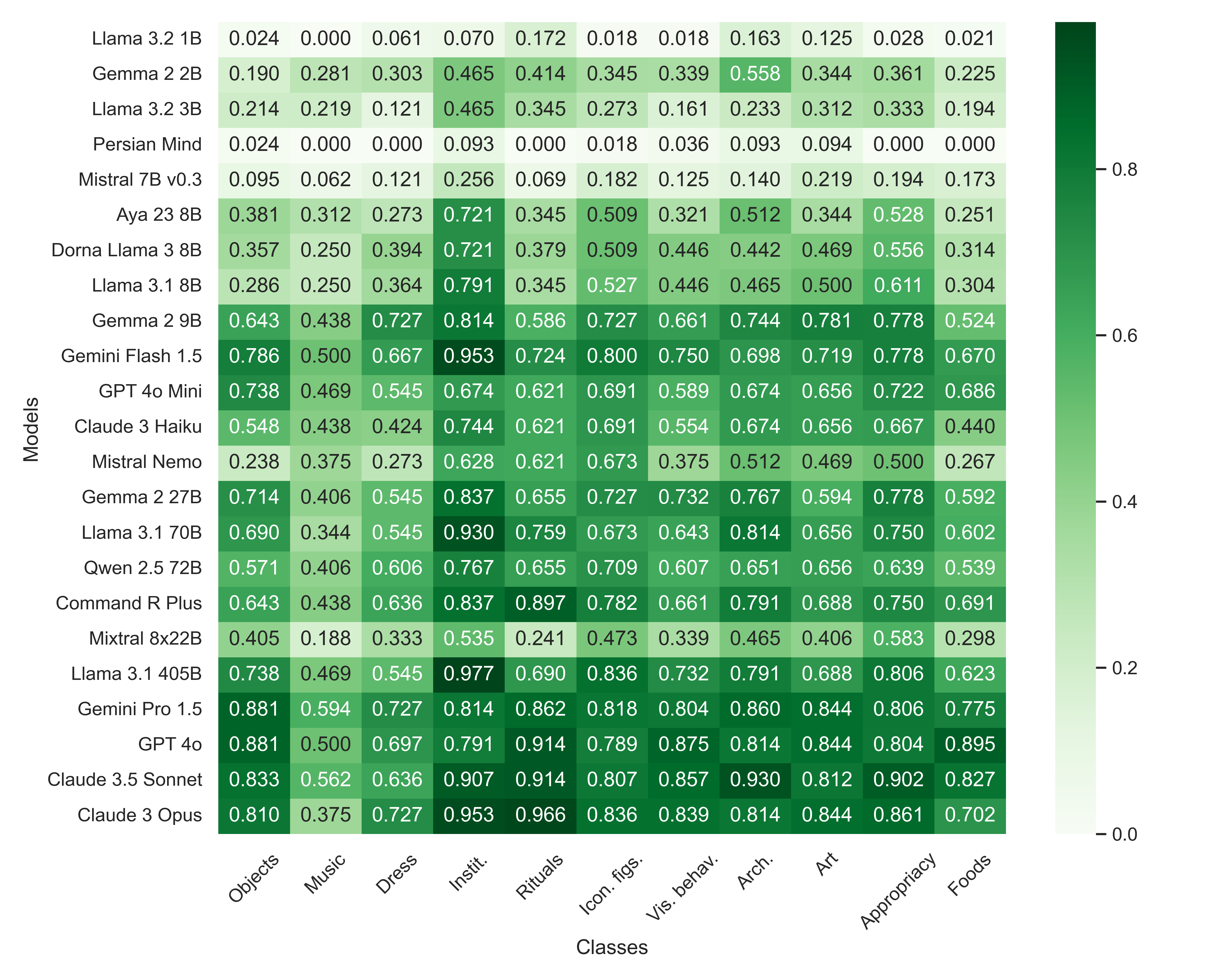}  % 
    \caption{The performance of different models across the 11 cultural categories in \textsc{\textsc{PerCul}}.}
    \label{fig:heatmap}
\end{figure*}

\paragraph{Accuracy and model size.}
Figure~\ref{fig:acc_experiment} shows the performance variation of models with their size (in terms of parameters).
As can be seen, there is a clear positive correlation between the number of parameters and the accuracy of models within each model family.
However, this relationship does not hold across different families. 
For instance, LLaMa 3.1 405B, while being the top model in the LLaMa family, its performance is close to Command-R-Plus, despite being nearly four times larger. 
Similarly, Gemma 2 9B's, which is comparable in accuracy to models 10x and 40x larger in terms of parameter count. These differences imply that architectural advancements and data efficiency might have a more substantial impact than mere size.
As Persian was not the target language for any of these multilingual LLMs, this outcome is expected due to the unknown quantity and quality of Persian data in their training set. 

\begin{figure}[t!]
    \centering
    \includegraphics[width=\columnwidth]{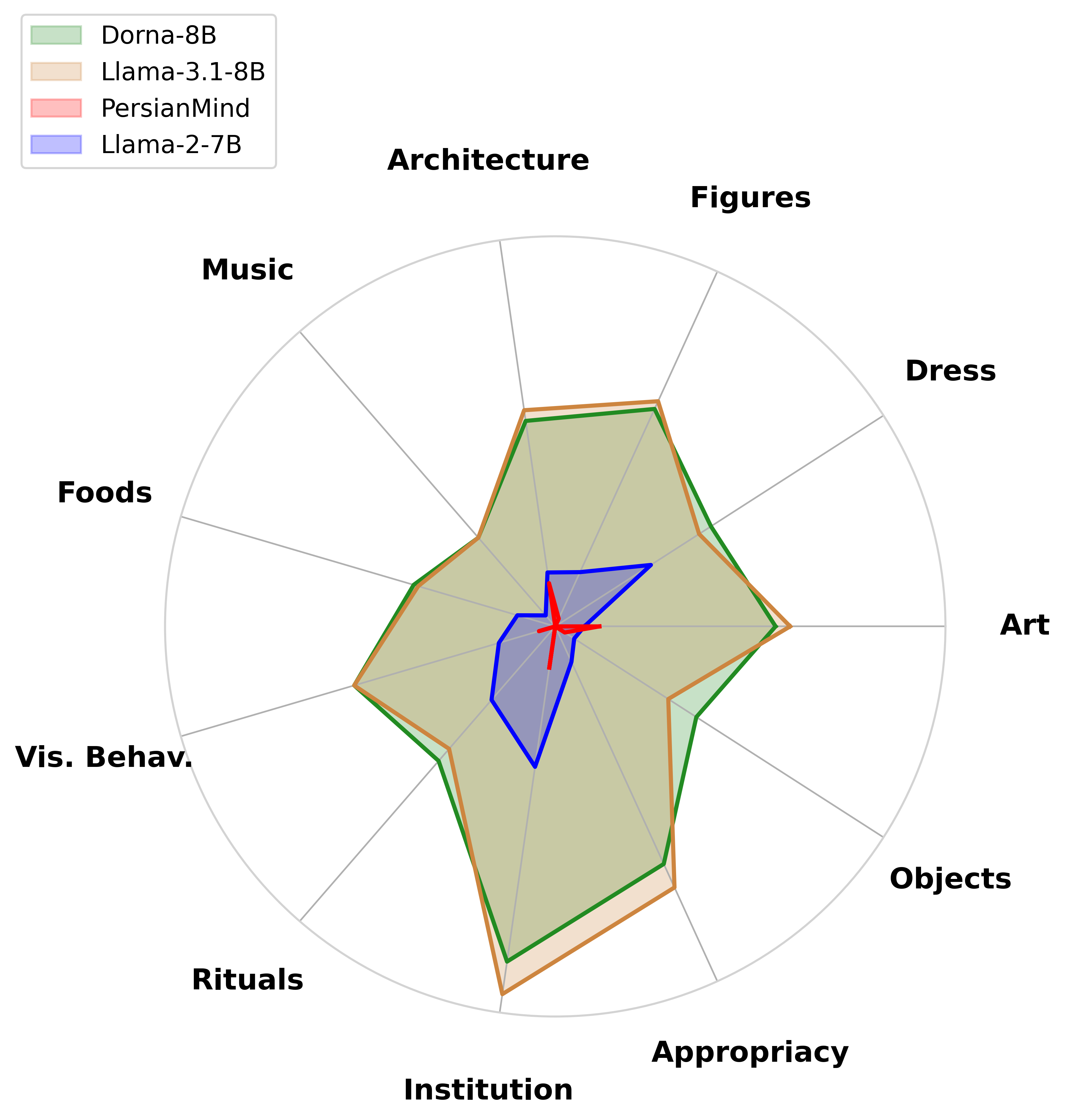}
    \caption{The impact of fine-tuning on Persian-specific models (both these models are the fine-tuned versions of Llama models).}
    \label{fig:fine_tuning}
\end{figure}

\paragraph{Persian (fine-tuned) models.}
Another interesting observation, as depicted in Figure~\ref{fig:fine_tuning}, is the negative impact of fine-tuning on Persian models. Both Persian models exhibited lower performance compared to their corresponding base models. Although it's fair to note that PersianMind refused to answer most of the questions by stating ''The answer is not available in the provided options''.
This could be attributed to the quality of the fine-tuning data which may have introduced noise or caused overfitting, resulting in a decline in the models' ability to generalize effectively. 
This finding prompts further investigation into the quality and relevance of the fine-tuning data, which we propose as a direction for future research. 

\subsection{Accuracy per category}
Figure~\ref{fig:heatmap} shows performance of the models across the 11 cultural categories.
Among these, music proves to be the most challenging, with the respective accuracies of 56.2\% and 59.4\% for 3.5-Sonnet and the best performing model on that category (Gemini Pro 1.5).
In contrast, rituals is the least difficult, with three models crossing over the 90\% performance (3.5 Sonnet, GPT 4o, and 3 Opus).
In general, we find that models from the same family exhibit higher performance correlations ($0.895 \pm 0.186$) compared to correlations between models from different providers ($0.576 \pm 0.223$).
Also, there is a very high correlation between the accuracy of models in each category and the overall performance over all categories (0.94 or above for all categories).

\subsection{Impact of Translation}
Since Persian is not a target language for most of these multilingual LLMs and their training corpora is predominantly English-based, one might assume that translating our stories into English could enhance these models' performances. 
We investigate the impact of translation on model performance to determine if they rely on translation as a proxy for understanding or have directly learned the concepts in the target language. 
To ensure the translation quality, we experiment with both Google Translate API and GPT 4o.
After careful investigation, we opted for GPT 4o given that it provided higher quality translations from Persian to English. 
\begin{figure}[t!]
    \centering
    \includegraphics[width=0.7\linewidth]{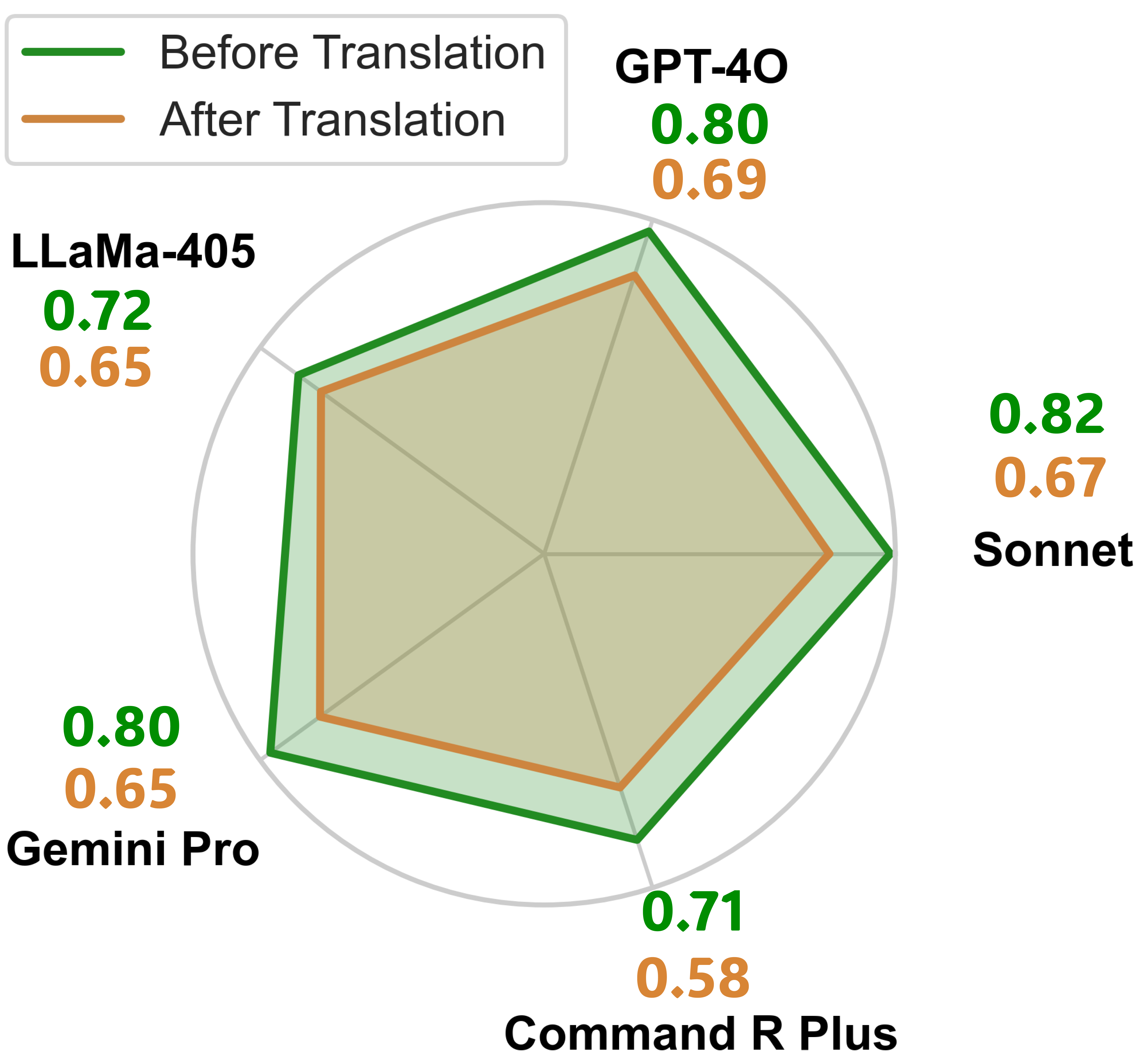}
    \caption{The degradation of best-performing models after dataset translation.}
    \label{fig:translation_effect}
\end{figure}
Figure~\ref{fig:translation_effect} displays the results for the best model in top-performing families, as evaluated using the English translation of the dataset.
As can be seen, the accuracy of these models decreases by 6.6\% to 14.5\% on the translated dataset.
To delve deeper into the reasons for this decline, we manually examine the results of Sonnet 3.5 (the best performing model) in both the original Persian and the translated samples. 
Having two sets of answers for these models, let's denote the set of correct answers in Persian as $P$ and that for English as $E$. 
Then, $P-E$ represents a set of answers where the model initially provided the correct answer but failed when the question was translated. 
To investigate the cause, we categorized the items in this set into three classes:
\begin{itemize}
    \item Nearly 19\% of the samples are correctly translated, but cultural nuances are lost in the process. For instance, the concept of \textit{respecting bread} holds significant meaning in Persian culture, but there is no direct equivalent in Western cultures, leading to loss in the benchmarking.
    
    \item Approximately 27\% of the samples encounter translation errors due to the lack of cultural equivalents in Western context. A notable example is the Persian culture's specific terminology for various bowls used for sugar powder, sugar cubes, and nabat (Persian crystal sugar). In translation, these distinct terms are generalized as a single ``sugar bowl,'' failing to capture the cultural specificity of the original text.
    
    \item The remaining 54\% of the samples, despite being accurately translated into English, are answered incorrectly by the model.
\end{itemize}
Conversely, we identify a smaller set $E-P$ where the translated samples are answered correctly by the model, while the original samples are not. These discrepancies can be attributed to the additional contextual information provided during the translation process. For instance, the term \textit{Tombak} is translated to \textit{Tombak (a type of Persian drum)}, or \textit{Abgoosht} as \textit{Abgoosht (Persian lamb stew)}. The inclusion of these descriptive phrases in the translation offers valuable clues that enable the LLMs to infer the correct answers more easily.

\subsection{Distraction Analysis}
To gain deeper insights into common failure patterns of LLMs concerning their understanding of Persian culture, we examine the distraction choices in \textsc{PerCul} and their success in deceiving the models. For this analysis, we evaluate how heuristic rules are distributed within each category (refer to Appendix~\ref{appendix:full_bench} for full model distributions) and consider how often a rule is chosen for its category as a measure of its effectiveness.

The effectiveness, distribution of heuristic rules within each category, of distractor options created by different heuristic rules in misleading models over different cultural categories.

\begin{figure}[!t]
    \centering
    \includegraphics[width=1\linewidth]{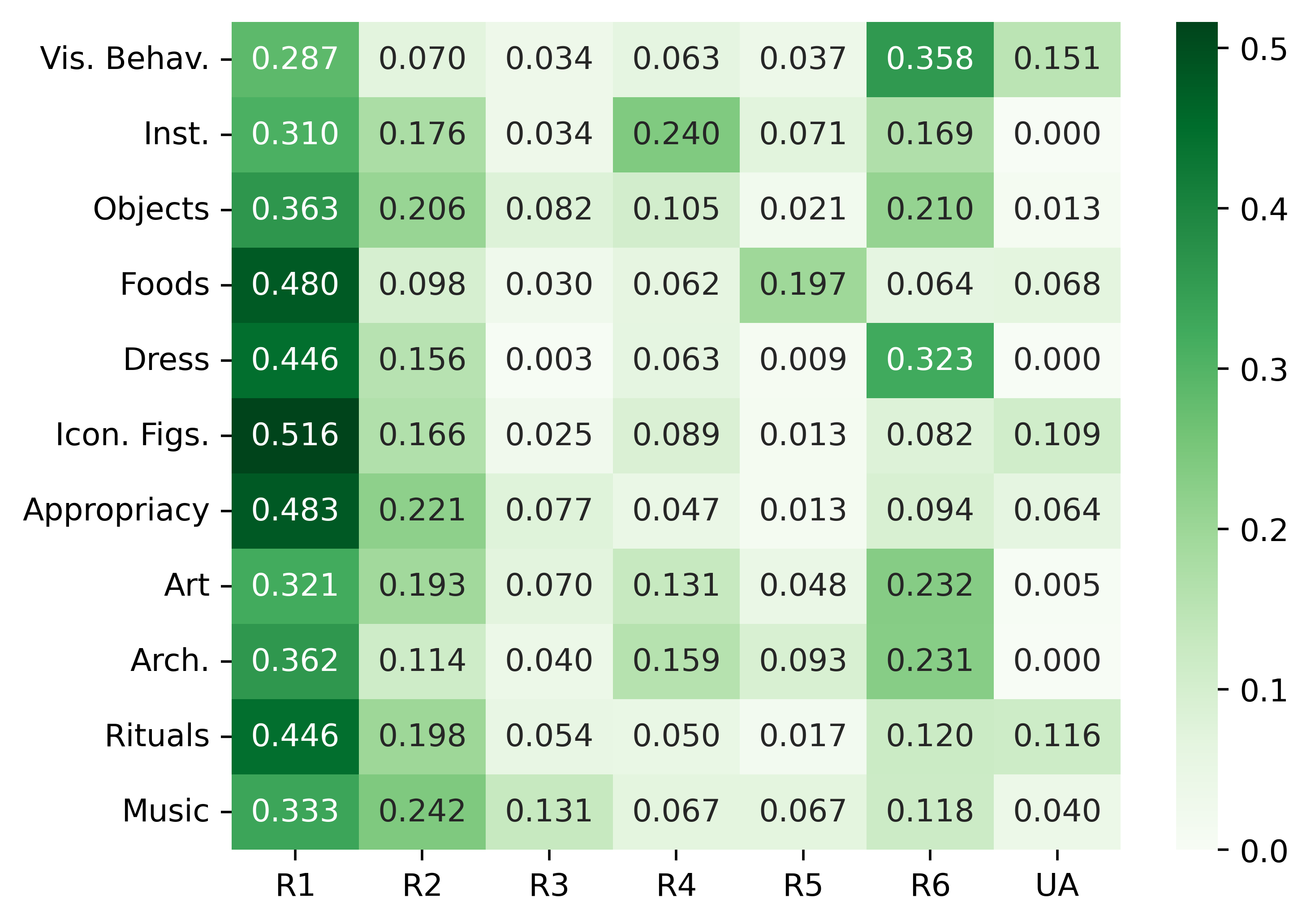}
    \caption{The impact and distribution of heuristic rules within each category, representing the effectiveness of each of them in misleading models. \textit{UA} represent options with complete rewrote by annotators.}
    \label{fig:heat_distraction}
\end{figure}
Our analysis across cultural categories (Figure~\ref{fig:heat_distraction}) shows that heuristic rule 1 (Partial Correctness) was consistently the most effective in misleading models. R1 creates options that are either partially correct or contain elements from the story but are ultimately incorrect. Its effectiveness stems from the fact that models often rely on surface-level semantic similarities rather than deeper cultural implications. When faced with partially correct yet incomplete information, models frequently select these seemingly plausible but incorrect options.
To illustrate this, consider the following example story from \textit{PerCul}:
\begin{quote} On a warm summer night, the people of a small village by the sea had gathered together. The sound of drums and various instruments was in the air, and everyone was dancing together. An old man with a white beard and eyes full of memories was sitting in a corner, smiling at the young people who were enthusiastically responding to the sound of music. Children were running happily among the crowd, and women were dancing beautifully in their colorful dresses. From time to time, the sound of a flute could be heard, giving the crowd a special atmosphere. These celebrations were always an excuse to get together and have fun, and no one wanted these beautiful moments to end. \end{quote}
In this example, models were asked to identify what cultural concept in the Music category is implied. The correct answer is ``Bandari music'', a distinctive genre of Persian music and dance, which is traditionally associated with the southern coastal regions of Iran. The presented distractor options to models are:
\begin{itemize} 
    \item R1 (Partial Correctness): ``Traditional coastal dance'' 
    \item R2 (Misinterpretation): ``Rural wedding celebration''
    \item R3 (Unrelated Fact): ``Any type of Persian music''
\end{itemize}

Models selecting R1 recognized the coastal setting and dancing but failed to connect it specifically to ``Bandari music,'' instead offering a partial surface-level response. R2, which models selected by misinterpreting the question, focused on the event rather than the cultural element. These findings highlight a limitation in LLMs: they often rely on surface-level details rather than synthesizing contextual clues to identify specific cultural traditions. This pattern is consistent across all 11 cultural categories and underscores a broader challenge in cultural understanding for current models. Additional examples are provided in Appendix~\ref{appendix:distractor_examples}.

\section{Conclusion}
In this paper, we introduced \textsc{PerCul}, a carefully curated dataset designed to assess LLMs' sensitivity towards Persian culture. 
Our dataset is non-trivial, as it employs implied concepts within conversations or story scenarios, rendering translation ineffective for solving our benchmark. 
The experiments demonstrated a significant performance gap between open-weight and closed-source LLMs for Persian culture. 
We also showed that the knowledge of Persian culture in LLMs is not dependent on the number of parameters when comparing inter-family models, whereas parameter count plays a crucial role in intra-family models. 
Lastly, our experiments revealed that current state-of-the-art Persian-specific LLMs fall short and even degrade in performance, when compared to their original base models, emphasizing the need for more effective methods, models, and higher-quality datasets to train Persian-specific LLMs. For future research, we suggest studying LLMs based on the final level of culture, namely \textit{informal} level, where categories are more subjective. One potential approach could involve assigning personalities to each LLM and observing their behavior in a simulated environment to evaluate the third level of cultural understanding.

\section*{Limitations}
During our research, we aimed to include annotators from diverse backgrounds and cities, but the majority were university students, which may introduce bias towards the Persian academic community and potentially limit the cultural knowledge captured in the dataset. Due to the inability to host most state-of-the-art LLMs locally, we relied on APIs to benchmark these models, restricting us to a specific set of models. While we managed to benchmark many SOTA models, the list is not exhaustive. Despite our efforts to encompass various aspects of Persian culture, there remain untapped areas such as individualism and communication that are not addressed in this work. These informal aspects of culture are inherently subjective and are hard to capture in the medium of text.

\section*{Ethics Statement}
This work presents various aspects of Persian culture through illustrative situations. While these aspects and their examples are gathered by a diverse group of Persian annotators and validated by another group, adhering to a carefully crafted manifesto, it is not entirely free from bias. Some sections of the dataset, particularly those concerning social norms and behaviors, contain information that mirrors the current state of Persian culture, regardless of whether it is unpleasant or criticized by new social movements. We included such content for the sake of comprehensiveness, and it does not necessarily reflect the authors' opinions on these matters.

\section*{Acknowledgments}
This research was in part supported by a grant from the School of Computer Science,
Institute for Research in Fundamental Sciences, IPM, Iran (No. CS1403-4-05).

\bibliography{custom}

\begin{thebibliography}{41}
\providecommand{\natexlab}[1]{#1}

\bibitem[{Abadani et~al.(2021)Abadani, Mozafari, Fatemi, Nematbakhsh, and Kazemi}]{parsquad_2021}
Negin Abadani, Jamshid Mozafari, Afsaneh Fatemi, Mohammd~Ali Nematbakhsh, and Arefeh Kazemi. 2021.
\newblock \href {https://doi.org/10.1109/ICWR51868.2021.9443126} {Parsquad: Machine translated squad dataset for persian question answering}.
\newblock In \emph{2021 7th International Conference on Web Research (ICWR)}, pages 163--168.

\bibitem[{Abaskohi et~al.(2024)Abaskohi, Baruni, Masoudi, Abbasi, Babalou, Edalat, Kamahi, Mahdizadeh~Sani, Naghavian, Namazifard, Sadeghi, and Yaghoobzadeh}]{persian_bench_2024}
Amirhossein Abaskohi, Sara Baruni, Mostafa Masoudi, Nesa Abbasi, Mohammad~Hadi Babalou, Ali Edalat, Sepehr Kamahi, Samin Mahdizadeh~Sani, Nikoo Naghavian, Danial Namazifard, Pouya Sadeghi, and Yadollah Yaghoobzadeh. 2024.
\newblock \href {https://aclanthology.org/2024.lrec-main.197} {Benchmarking large language models for {P}ersian: A preliminary study focusing on {C}hat{GPT}}.
\newblock In \emph{Proceedings of the 2024 Joint International Conference on Computational Linguistics, Language Resources and Evaluation (LREC-COLING 2024)}, pages 2189--2203, Torino, Italia. ELRA and ICCL.

\bibitem[{AlKhamissi et~al.(2024)AlKhamissi, ElNokrashy, Alkhamissi, and Diab}]{cultural_us_egypt_2024}
Badr AlKhamissi, Muhammad ElNokrashy, Mai Alkhamissi, and Mona Diab. 2024.
\newblock \href {https://doi.org/10.18653/v1/2024.acl-long.671} {Investigating cultural alignment of large language models}.
\newblock In \emph{Proceedings of the 62nd Annual Meeting of the Association for Computational Linguistics (Volume 1: Long Papers)}, pages 12404--12422, Bangkok, Thailand. Association for Computational Linguistics.

\bibitem[{Amirkhani et~al.(2023)Amirkhani, AzariJafari, Faridan-Jahromi, Kouhkan, Pourjafari, and Amirak}]{farstail_2023}
Hossein Amirkhani, Mohammad AzariJafari, Soroush Faridan-Jahromi, Zeinab Kouhkan, Zohreh Pourjafari, and Azadeh Amirak. 2023.
\newblock \href {https://doi.org/10.1007/s00500-023-08959-3} {Farstail: a persian natural language inference dataset}.
\newblock \emph{Soft Computing}.

\bibitem[{Anthropic(2024)}]{model_sonnet}
Anthropic. 2024.
\newblock \href {https://www-cdn.anthropic.com/fed9cc193a14b84131812372d8d5857f8f304c52/Model_Card_Claude_3_Addendum.pdf} {Claude 3.5 sonnet model card addendum}.

\bibitem[{Bhatt and Diaz(2024)}]{intro_2024_0}
Shaily Bhatt and Fernando Diaz. 2024.
\newblock \href {https://arxiv.org/abs/2406.11565} {Extrinsic evaluation of cultural competence in large language models}.
\newblock \emph{Preprint}, arXiv:2406.11565.

\bibitem[{Chang et~al.(2024)Chang, Wang, Wang, Wu, Yang, Zhu, Chen, Yi, Wang, Wang, Ye, Zhang, Chang, Yu, Yang, and Xie}]{llm_evaluation}
Yupeng Chang, Xu~Wang, Jindong Wang, Yuan Wu, Linyi Yang, Kaijie Zhu, Hao Chen, Xiaoyuan Yi, Cunxiang Wang, Yidong Wang, Wei Ye, Yue Zhang, Yi~Chang, Philip~S. Yu, Qiang Yang, and Xing Xie. 2024.
\newblock \href {https://doi.org/10.1145/3641289} {A survey on evaluation of large language models}.
\newblock \emph{ACM Trans. Intell. Syst. Technol.}, 15(3).

\bibitem[{Chiu et~al.(2024)Chiu, Jiang, Lin, Park, Li, Ravi, Bhatia, Antoniak, Tsvetkov, Shwartz, and Choi}]{cultural_en_5_2024}
Yu~Ying Chiu, Liwei Jiang, Bill~Yuchen Lin, Chan~Young Park, Shuyue~Stella Li, Sahithya Ravi, Mehar Bhatia, Maria Antoniak, Yulia Tsvetkov, Vered Shwartz, and Yejin Choi. 2024.
\newblock \href {https://arxiv.org/abs/2410.02677} {Culturalbench: a robust, diverse and challenging benchmark on measuring the (lack of) cultural knowledge of llms}.
\newblock \emph{Preprint}, arXiv:2410.02677.

\bibitem[{Darvishi et~al.(2023)Darvishi, Shahbodaghkhan, Abbasiantaeb, and Momtazi}]{pquad_2023}
Kasra Darvishi, Newsha Shahbodaghkhan, Zahra Abbasiantaeb, and Saeedeh Momtazi. 2023.
\newblock \href {https://doi.org/10.1016/j.csl.2023.101486} {Pquad: A persian question answering dataset}.
\newblock \emph{Computer Speech and Language}, 80:101486.

\bibitem[{Dubey et~al.(2024)Dubey, Jauhri, Pandey, Kadian, Al-Dahle, Letman, Mathur, Schelten, Yang, Fan et~al.}]{llama3}
Abhimanyu Dubey, Abhinav Jauhri, Abhinav Pandey, Abhishek Kadian, Ahmad Al-Dahle, Aiesha Letman, Akhil Mathur, Alan Schelten, Amy Yang, Angela Fan, et~al. 2024.
\newblock The llama 3 herd of models.
\newblock \emph{arXiv preprint arXiv:2407.21783}.

\bibitem[{Fung et~al.(2024)Fung, Zhao, Doo, Sun, and Ji}]{cultural_multi_1_2024}
Yi~Fung, Ruining Zhao, Jae Doo, Chenkai Sun, and Heng Ji. 2024.
\newblock \href {https://arxiv.org/abs/2402.09369} {Massively multi-cultural knowledge acquisition and lm benchmarking}.
\newblock \emph{Preprint}, arXiv:2402.09369.

\bibitem[{Ghahroodi et~al.(2024)Ghahroodi, Nouri, Sanian, Sahebi, Dastgheib, Asgari, Baghshah, and Rohban}]{khayyam_2024}
Omid Ghahroodi, Marzia Nouri, Mohammad~Vali Sanian, Alireza Sahebi, Doratossadat Dastgheib, Ehsaneddin Asgari, Mahdieh~Soleymani Baghshah, and Mohammad~Hossein Rohban. 2024.
\newblock \href {https://arxiv.org/abs/2404.06644} {Khayyam challenge (persianmmlu): Is your llm truly wise to the persian language?}
\newblock \emph{Preprint}, arXiv:2404.06644.

\bibitem[{Huang et~al.(2024)Huang, Yu, Zhu, Sun, Cheng, Dingjie, Chen, Alharthi, An, He, Liu, Chen, Li, Wang, Zhang, Sun, Wan, Li, and Xu}]{acegpt_2024}
Huang Huang, Fei Yu, Jianqing Zhu, Xuening Sun, Hao Cheng, Song Dingjie, Zhihong Chen, Mosen Alharthi, Bang An, Juncai He, Ziche Liu, Junying Chen, Jianquan Li, Benyou Wang, Lian Zhang, Ruoyu Sun, Xiang Wan, Haizhou Li, and Jinchao Xu. 2024.
\newblock \href {https://doi.org/10.18653/v1/2024.naacl-long.450} {{A}ce{GPT}, localizing large language models in {A}rabic}.
\newblock In \emph{Proceedings of the 2024 Conference of the North American Chapter of the Association for Computational Linguistics: Human Language Technologies (Volume 1: Long Papers)}, pages 8139--8163, Mexico City, Mexico. Association for Computational Linguistics.

\bibitem[{Katan and Taibi(2021)}]{katan}
David Katan and Mustapha Taibi. 2021.
\newblock \href {https://www.amazon.com/Translating-Cultures-David-Katan/dp/113834446X} {\emph{Translating Cultures: An Introduction for Translators, Interpreters and Mediators, Third Edition}}.

\bibitem[{Khashabi et~al.(2021)Khashabi, Cohan, Shakeri, Hosseini, Pezeshkpour, Alikhani, Aminnaseri, Bitaab, Brahman, Ghazarian, Gheini, Kabiri, Mahabagdi, Memarrast, Mosallanezhad, Noury, Raji, Rasooli, Sadeghi, Azer, Samghabadi, Shafaei, Sheybani, Tazarv, and Yaghoobzadeh}]{parsinlu_2021}
Daniel Khashabi, Arman Cohan, Siamak Shakeri, Pedram Hosseini, Pouya Pezeshkpour, Malihe Alikhani, Moin Aminnaseri, Marzieh Bitaab, Faeze Brahman, Sarik Ghazarian, Mozhdeh Gheini, Arman Kabiri, Rabeeh~Karimi Mahabagdi, Omid Memarrast, Ahmadreza Mosallanezhad, Erfan Noury, Shahab Raji, Mohammad~Sadegh Rasooli, Sepideh Sadeghi, Erfan~Sadeqi Azer, Niloofar~Safi Samghabadi, Mahsa Shafaei, Saber Sheybani, Ali Tazarv, and Yadollah Yaghoobzadeh. 2021.
\newblock {ParsiNLU: A Suite of Language Understanding Challenges for Persian}.
\newblock \emph{Transactions of the Association for Computational Linguistics}, 9:1147--1162.

\bibitem[{Kim et~al.(2024)Kim, Suk, Oh, Yoo, Thorne, and Oh}]{cultural_korean_2024}
Eunsu Kim, Juyoung Suk, Philhoon Oh, Haneul Yoo, James Thorne, and Alice Oh. 2024.
\newblock \href {https://aclanthology.org/2024.lrec-main.296} {{CLI}c{K}: A benchmark dataset of cultural and linguistic intelligence in {K}orean}.
\newblock In \emph{Proceedings of the 2024 Joint International Conference on Computational Linguistics, Language Resources and Evaluation (LREC-COLING 2024)}, pages 3335--3346, Torino, Italia. ELRA and ICCL.

\bibitem[{Li et~al.(2024)Li, Chen, Wang, Sitaram, and Xie}]{intro_2024_1}
Cheng Li, Mengzhou Chen, Jindong Wang, Sunayana Sitaram, and Xing Xie. 2024.
\newblock \href {https://arxiv.org/abs/2402.10946} {Culturellm: Incorporating cultural differences into large language models}.
\newblock \emph{Preprint}, arXiv:2402.10946.

\bibitem[{Manrai et~al.(2019)Manrai, Manrai, Lascu, and Friedeborn}]{halls_ref_2019}
Lalita~A. Manrai, Ajay~K. Manrai, Dana-Nicoleta Lascu, and Stefanie Friedeborn. 2019.
\newblock \href {https://api.semanticscholar.org/CorpusID:149838700} {Determinants and effects of cultural context: A review, conceptual model, and propositions}.
\newblock \emph{Journal of Global Marketing}, 32:67 -- 82.

\bibitem[{Masoud et~al.(2023)Masoud, Liu, Ferianc, Treleaven, and Rodrigues}]{cultural_en_3_2023}
Reem~I. Masoud, Ziquan Liu, Martin Ferianc, Philip Treleaven, and Miguel Rodrigues. 2023.
\newblock \href {https://arxiv.org/abs/2309.12342} {Cultural alignment in large language models: An explanatory analysis based on hofstede's cultural dimensions}.
\newblock \emph{Preprint}, arXiv:2309.12342.

\bibitem[{Myung et~al.(2024)Myung, Lee, Zhou, Jin, Putri, Antypas, Borkakoty, Kim, P{\'e}rez-Almendros, Ayele, Guti'errez-Basulto, Ib'anez-Garc'ia, Lee, Muhammad, Park, Rzayev, White, Yimam, Pilehvar, Ousidhoum, Camacho-Collados, and Oh}]{blend_2024}
Jun-Hee Myung, Nayeon Lee, Yi~Zhou, Jiho Jin, Rifki~Afina Putri, Dimosthenis Antypas, Hsuvas Borkakoty, Eunsu Kim, Carla P{\'e}rez-Almendros, Abinew~Ali Ayele, V'ictor Guti'errez-Basulto, Yazm'in Ib'anez-Garc'ia, Hwaran Lee, Shamsuddeen~Hassan Muhammad, Kiwoong Park, Anar Rzayev, Nina White, Seid~Muhie Yimam, Mohammad~Taher Pilehvar, Nedjma~Djouhra Ousidhoum, Jos{\'e} Camacho-Collados, and Alice Oh. 2024.
\newblock \href {https://api.semanticscholar.org/CorpusID:270521296} {Blend: A benchmark for llms on everyday knowledge in diverse cultures and languages}.
\newblock \emph{ArXiv}, abs/2406.09948.

\bibitem[{Naous et~al.(2024)Naous, Ryan, Ritter, and Xu}]{beer_2024}
Tarek Naous, Michael Ryan, Alan Ritter, and Wei Xu. 2024.
\newblock \href {https://doi.org/10.18653/v1/2024.acl-long.862} {Having beer after prayer? measuring cultural bias in large language models}.
\newblock In \emph{Proceedings of the 62nd Annual Meeting of the Association for Computational Linguistics (Volume 1: Long Papers)}, pages 16366--16393, Bangkok, Thailand. Association for Computational Linguistics.

\bibitem[{Noorbakhsh et~al.(2021)Noorbakhsh, Sulaiman, Sharifi, Roy, and Jamshidi}]{trans_2021}
Kimia Noorbakhsh, Modar Sulaiman, Mahdi Sharifi, Kallol Roy, and Pooyan Jamshidi. 2021.
\newblock \href {https://api.semanticscholar.org/CorpusID:238419670} {Pretrained language models are symbolic mathematics solvers too!}
\newblock \emph{ArXiv}, abs/2110.03501.

\bibitem[{PartAI(2024)}]{dorna}
PartAI. 2024.
\newblock {P}art{A}{I}/{D}orna-{L}lama3-8{B}-{I}nstruct.
\newblock \url{https://huggingface.co/PartAI/Dorna-Llama3-8B-Instruct}.
\newblock [Accessed 13-10-2024].

\bibitem[{Qin et~al.(2024)Qin, Song, Hu, Yao, Cho, Wang, Wu, Liu, Liu, and Yu}]{llm_bench_6_if}
Yiwei Qin, Kaiqiang Song, Yebowen Hu, Wenlin Yao, Sangwoo Cho, Xiaoyang Wang, Xuansheng Wu, Fei Liu, Pengfei Liu, and Dong Yu. 2024.
\newblock \href {https://doi.org/10.18653/v1/2024.findings-acl.772} {{I}n{F}o{B}ench: Evaluating instruction following ability in large language models}.
\newblock In \emph{Findings of the Association for Computational Linguistics ACL 2024}, pages 13025--13048, Bangkok, Thailand and virtual meeting. Association for Computational Linguistics.

\bibitem[{Rao et~al.(2024)Rao, Yerukola, Shah, Reinecke, and Sap}]{cultural_en_4_2024}
Abhinav Rao, Akhila Yerukola, Vishwa Shah, Katharina Reinecke, and Maarten Sap. 2024.
\newblock \href {https://arxiv.org/abs/2404.12464} {Normad: A benchmark for measuring the cultural adaptability of large language models}.
\newblock \emph{Preprint}, arXiv:2404.12464.

\bibitem[{Rein et~al.(2023)Rein, Hou, Stickland, Petty, Pang, Dirani, Michael, and Bowman}]{llm_bench_3_k}
David Rein, Betty~Li Hou, Asa~Cooper Stickland, Jackson Petty, Richard~Yuanzhe Pang, Julien Dirani, Julian Michael, and Samuel~R. Bowman. 2023.
\newblock \href {https://arxiv.org/abs/2311.12022} {Gpqa: A graduate-level google-proof q\&a benchmark}.
\newblock \emph{Preprint}, arXiv:2311.12022.

\bibitem[{Rostami et~al.(2024)Rostami, Salemi, and Dousti}]{persianmind}
Pedram Rostami, Ali Salemi, and Mohammad~Javad Dousti. 2024.
\newblock \href {https://arxiv.org/abs/2401.06466} {Persianmind: A cross-lingual persian-english large language model}.
\newblock \emph{Preprint}, arXiv:2401.06466.

\bibitem[{Saffari et~al.(2024)Saffari, Shafiei, and Pierri}]{psn_2024}
Hamidreza Saffari, Mohammadamin Shafiei, and Francesco Pierri. 2024.
\newblock \href {https://arxiv.org/abs/2406.09123} {Psn: Persian social norms dataset for cross-cultural ai}.
\newblock \emph{Preprint}, arXiv:2406.09123.

\bibitem[{Singh et~al.(2024)Singh, Patidar, and Vig}]{halls_ref_2024}
Pushpdeep Singh, Mayur Patidar, and Lovekesh Vig. 2024.
\newblock \href {https://arxiv.org/abs/2406.14504} {Translating across cultures: Llms for intralingual cultural adaptation}.
\newblock \emph{Preprint}, arXiv:2406.14504.

\bibitem[{Sprague et~al.(2024)Sprague, Ye, Bostrom, Chaudhuri, and Durrett}]{llm_bench_4_rm}
Zayne Sprague, Xi~Ye, Kaj Bostrom, Swarat Chaudhuri, and Greg Durrett. 2024.
\newblock \href {https://arxiv.org/abs/2310.16049} {Musr: Testing the limits of chain-of-thought with multistep soft reasoning}.
\newblock \emph{Preprint}, arXiv:2310.16049.

\bibitem[{Suzgun et~al.(2023)Suzgun, Scales, Sch{\"a}rli, Gehrmann, Tay, Chung, Chowdhery, Le, Chi, Zhou, and Wei}]{llm_bench_1_rm}
Mirac Suzgun, Nathan Scales, Nathanael Sch{\"a}rli, Sebastian Gehrmann, Yi~Tay, Hyung~Won Chung, Aakanksha Chowdhery, Quoc Le, Ed~Chi, Denny Zhou, and Jason Wei. 2023.
\newblock \href {https://doi.org/10.18653/v1/2023.findings-acl.824} {Challenging {BIG}-bench tasks and whether chain-of-thought can solve them}.
\newblock In \emph{Findings of the Association for Computational Linguistics: ACL 2023}, pages 13003--13051, Toronto, Canada. Association for Computational Linguistics.

\bibitem[{Tao et~al.(2024)Tao, Viberg, Baker, and Kizilcec}]{intro_2024_3}
Yan Tao, Olga Viberg, Ryan~S Baker, and René~F Kizilcec. 2024.
\newblock \href {https://doi.org/10.1093/pnasnexus/pgae346} {{Cultural bias and cultural alignment of large language models}}.
\newblock \emph{PNAS Nexus}, 3(9):pgae346.

\bibitem[{Team(2024)}]{model_gpt4}
OpenAI Team. 2024.
\newblock \href {https://arxiv.org/abs/2303.08774} {Gpt-4 technical report}.
\newblock \emph{Preprint}, arXiv:2303.08774.

\bibitem[{Tedlock and Mannheim(1995)}]{culture_in_dialogue}
Dennis Tedlock and Bruce Mannheim. 1995.
\newblock \emph{The dialogic emergence of culture}.
\newblock University of Illinois Press.

\bibitem[{Thier(2013)}]{halls_ref_2013}
Michael Thier. 2013.
\newblock \href {https://api.semanticscholar.org/CorpusID:268309397} {Cultural awareness logs: A method for increasing international-mindedness among high school and middle school students}.
\newblock \emph{English Journal}.

\bibitem[{Tikhonov et~al.(2021)Tikhonov, Samenko, and Yamshchikov}]{story_2021}
Alexey Tikhonov, Igor Samenko, and Ivan~P. Yamshchikov. 2021.
\newblock \href {https://doi.org/10.18653/v1/2021.eval4nlp-1.4} {{S}tory{DB}: Broad multi-language narrative dataset}.
\newblock In \emph{Proceedings of the 2nd Workshop on Evaluation and Comparison of NLP Systems}, pages 32--39, Punta Cana, Dominican Republic. Association for Computational Linguistics.

\bibitem[{Wang et~al.(2024{\natexlab{a}})Wang, Jiao, Huang, Dai, Huang, Tu, and Lyu}]{cultural_en_1_2024}
Wenxuan Wang, Wenxiang Jiao, Jingyuan Huang, Ruyi Dai, Jen-tse Huang, Zhaopeng Tu, and Michael Lyu. 2024{\natexlab{a}}.
\newblock \href {https://doi.org/10.18653/v1/2024.acl-long.345} {Not all countries celebrate thanksgiving: On the cultural dominance in large language models}.
\newblock In \emph{Proceedings of the 62nd Annual Meeting of the Association for Computational Linguistics (Volume 1: Long Papers)}, pages 6349--6384, Bangkok, Thailand. Association for Computational Linguistics.

\bibitem[{Wang et~al.(2024{\natexlab{b}})Wang, Ma, Zhang, Ni, Chandra, Guo, Ren, Arulraj, He, Jiang, Li, Ku, Wang, Zhuang, Fan, Yue, and Chen}]{llm_bench_5_k}
Yubo Wang, Xueguang Ma, Ge~Zhang, Yuansheng Ni, Abhranil Chandra, Shiguang Guo, Weiming Ren, Aaran Arulraj, Xuan He, Ziyan Jiang, Tianle Li, Max Ku, Kai Wang, Alex Zhuang, Rongqi Fan, Xiang Yue, and Wenhu Chen. 2024{\natexlab{b}}.
\newblock \href {https://arxiv.org/abs/2406.01574} {Mmlu-pro: A more robust and challenging multi-task language understanding benchmark}.
\newblock \emph{Preprint}, arXiv:2406.01574.

\bibitem[{Wang et~al.(2024{\natexlab{c}})Wang, Zhu, Kong, Wei, Yi, Xie, and Sang}]{cultural_en_2_2024}
Yuhang Wang, Yanxu Zhu, Chao Kong, Shuyu Wei, Xiaoyuan Yi, Xing Xie, and Jitao Sang. 2024{\natexlab{c}}.
\newblock \href {https://doi.org/10.18653/v1/2024.c3nlp-1.1} {{CDE}val: A benchmark for measuring the cultural dimensions of large language models}.
\newblock In \emph{Proceedings of the 2nd Workshop on Cross-Cultural Considerations in NLP}, pages 1--16, Bangkok, Thailand. Association for Computational Linguistics.

\bibitem[{Zhao et~al.(2024)Zhao, Zhang, Chen, Kawaguchi, and Bing}]{trans_2024}
Yiran Zhao, Wenxuan Zhang, Guizhen Chen, Kenji Kawaguchi, and Lidong Bing. 2024.
\newblock \href {https://api.semanticscholar.org/CorpusID:268063798} {How do large language models handle multilingualism?}
\newblock \emph{ArXiv}, abs/2402.18815.

\bibitem[{Zhou et~al.(2023)Zhou, Lu, Mishra, Brahma, Basu, Luan, Zhou, and Hou}]{llm_bench_0_if}
Jeffrey Zhou, Tianjian Lu, Swaroop Mishra, Siddhartha Brahma, Sujoy Basu, Yi~Luan, Denny Zhou, and Le~Hou. 2023.
\newblock \href {https://arxiv.org/abs/2311.07911} {Instruction-following evaluation for large language models}.
\newblock \emph{Preprint}, arXiv:2311.07911.

\end{thebibliography}

\clearpage
\appendix
\section{Annotators}
\label{appendix:annotators}
This section provides information about the eight Persian participants involved in the annotation process. The annotators vary in age (18-28 years), gender (3 female, 5 male), and come from different cities across Iran. Most participants are pursuing graduate studies, with one high school student and one doctoral candidate. Table~\ref{tab:demographics} presents the detailed demographic information of our annotators, including their sex, place of birth (PoB), age, and educational background.

\begin{table}[!t]
\centering
\resizebox{\linewidth}{!}{
\setlength{\tabcolsep}{12pt}
\begin{tabular}{llrl}
\toprule
Sex & PoB & Age & Education \\
\midrule
Female & Isfahan & 28 & PhD Student \\
Male & Yazd & 28 & MSc Student \\
Male & Kashan & 27 & MSc Student \\
Male & Shiraz & 27 & MSc Student \\
Male & Behbahan & 25 & MSc Student \\
Female & Karaj & 24 & MSc Student \\
Male & Shiraz & 24 & MSc Student \\
Male & Shiraz & 23 & MSc Student \\
Female & Tehran & 18 & Student \\
\bottomrule
\end{tabular}
}
\caption{Education and Demographic Data of Participants.}
\label{tab:demographics}
\end{table}

\section{Annotation Guidelines}
\label{appendix:guideline}

This section outlines the detailed guidelines provided to annotators for different phases of our data creation and evaluation process.

\subsection{Seed Topic Generation}
 This phase involves the creation of seed topics across 11 Persian cultural categories. These topics serve as foundational elements for story generation using large language models (LLMs). The following guidelines should be followed:
\begin{itemize}
\item Select topics that are broadly representative of Persian culture, avoiding those specific to regional subcultures
\item Choose topics with cultural distinctiveness, rather than universal or generic themes
\item Ensure each topic is unique and distinguishable from others in the dataset
\item Focus on enduring cultural elements that are neither too historical nor too contemporary
\end{itemize}

\subsection{Facet Identification}
This phase involves identifying and defining key facets within each cultural category. These facets serve as structured characteristics that describe each seed topic in detail. For example, when examining foods, facets might include preparation methods, cultural significance, and traditional serving contexts. The following guidelines should be followed:

\begin{itemize}
\item Define facets comprehensively to account for potential LLM knowledge gaps
\item Include distinctive features that can facilitate unique story generation
\item Maintain brevity and clarity in facet descriptions
\item Rely solely on human knowledge, LLM-generated content is not allowed
\end{itemize}

\subsection{Metadata Creation}
This phase involves creating detailed metadata to ground LLMs during story generation. This metadata serves as factual foundation to prevent hallucinations and ensure cultural accuracy in generated stories. Here are the guidelines to follow:
\begin{itemize}
    \item Ensure metadata differentiates seed topics within a category
    \item Provide sufficient clues to allow metadata inference from stories
    \item Avoid using LLMs for data generation
    \item Use Google sparingly and only when metadata cannot be filled with certainty
    \item Minimize reliance on Wikipedia due to LLM exposure
    \item Maintain precision and conciseness in metadata
\end{itemize}

\subsection{Distractor Choice Selection}
This phase involves evaluating and selecting distractor options generated by LLMs. The selection process uses six predefined heuristic rules to ensure quality and diversity of multiple-choice options. The following points should be considered:
\begin{itemize}
\item Try to choose distractors that represent different heuristic categories
\item Apply consistent selection criteria based on provided rule definitions
\item Ensure selected distractors are appropriate for Persian cultural knowledge
\item Avoid redundant or overlapping options in the final selection
\end{itemize}

\subsection{Human Baseline Generation}
This phase involves determining cultural concepts that are implied in short stories. Each story indirectly references a Persian cultural element, accompanied by a comprehension question and four answer options targeting the implied concept. The following guidelines should be followed:
\begin{itemize}
\item Base your answer on personal knowledge without LLM assistance
\item Limit internet research to essential fact verification
\item Select the most precise option when multiple choices appear partially correct
\end{itemize}

\section{Prompts}
\label{appendix:prompts}
\begin{figure}[!h]
    \centering
    \includegraphics[width=1\linewidth]{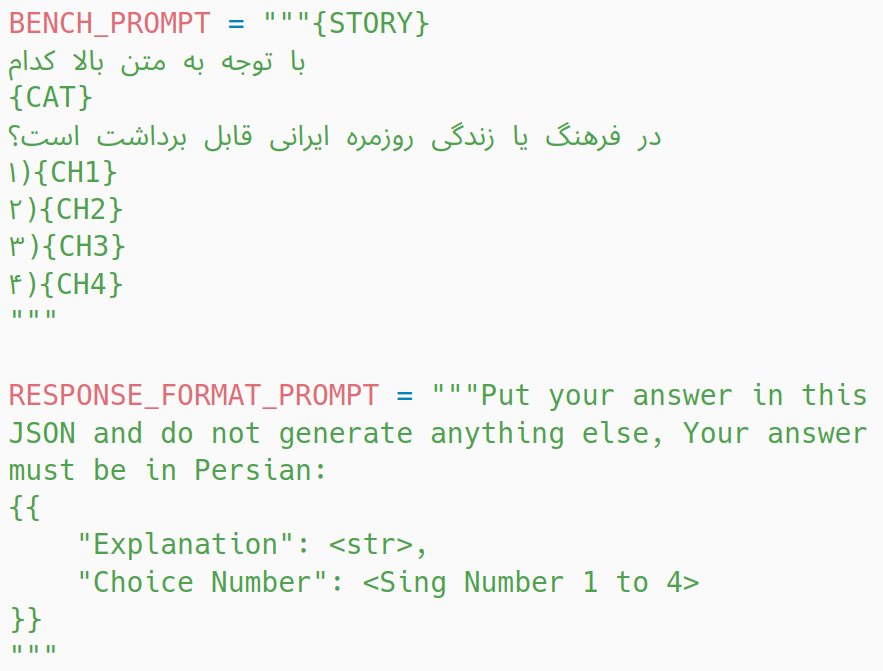}
    \caption{Prompt used for benchmarking different LLMs.}
    \label{fig:prompt_benchmarking}
\end{figure}

\begin{figure}[!h]
    \centering
    \includegraphics[width=1\linewidth]{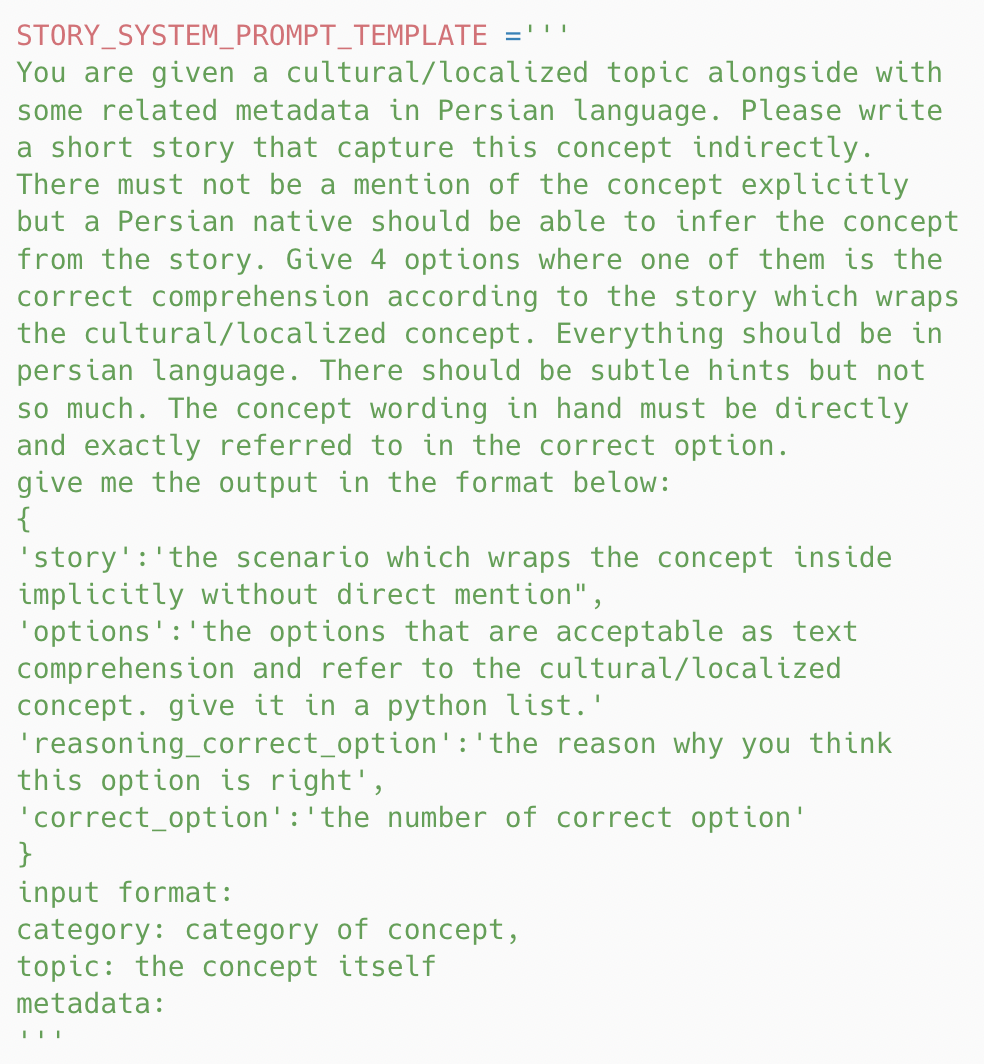}
    \caption{Prompt used for Story Generation.}
    \label{fig:prompt_story_generation}
\end{figure}

\begin{figure}[!h]
    \centering
    \includegraphics[width=1\linewidth]{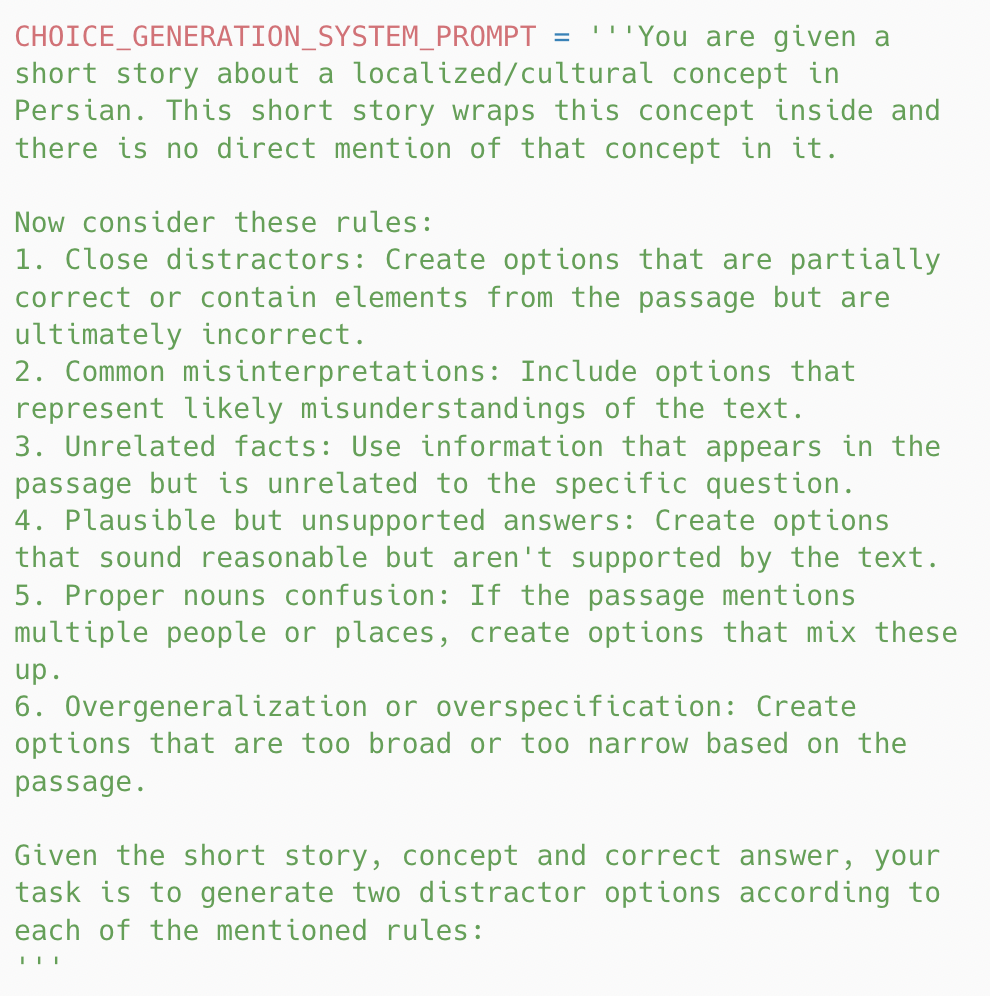}
    \caption{Prompt used for distractor generation.}
    \label{fig:prompt_choice_generation}
\end{figure}

In this section, we have prompts that was used for LLM generation and benchmarking in different steps of the work shown in Figure~\ref{fig:prompt_benchmarking}, Figure ~\ref{fig:prompt_story_generation} and Figure~\ref{fig:prompt_choice_generation}.

\newpage
\section{User Interfaces}
\begin{figure}[!h]
    \centering
    \includegraphics[width=1\linewidth]{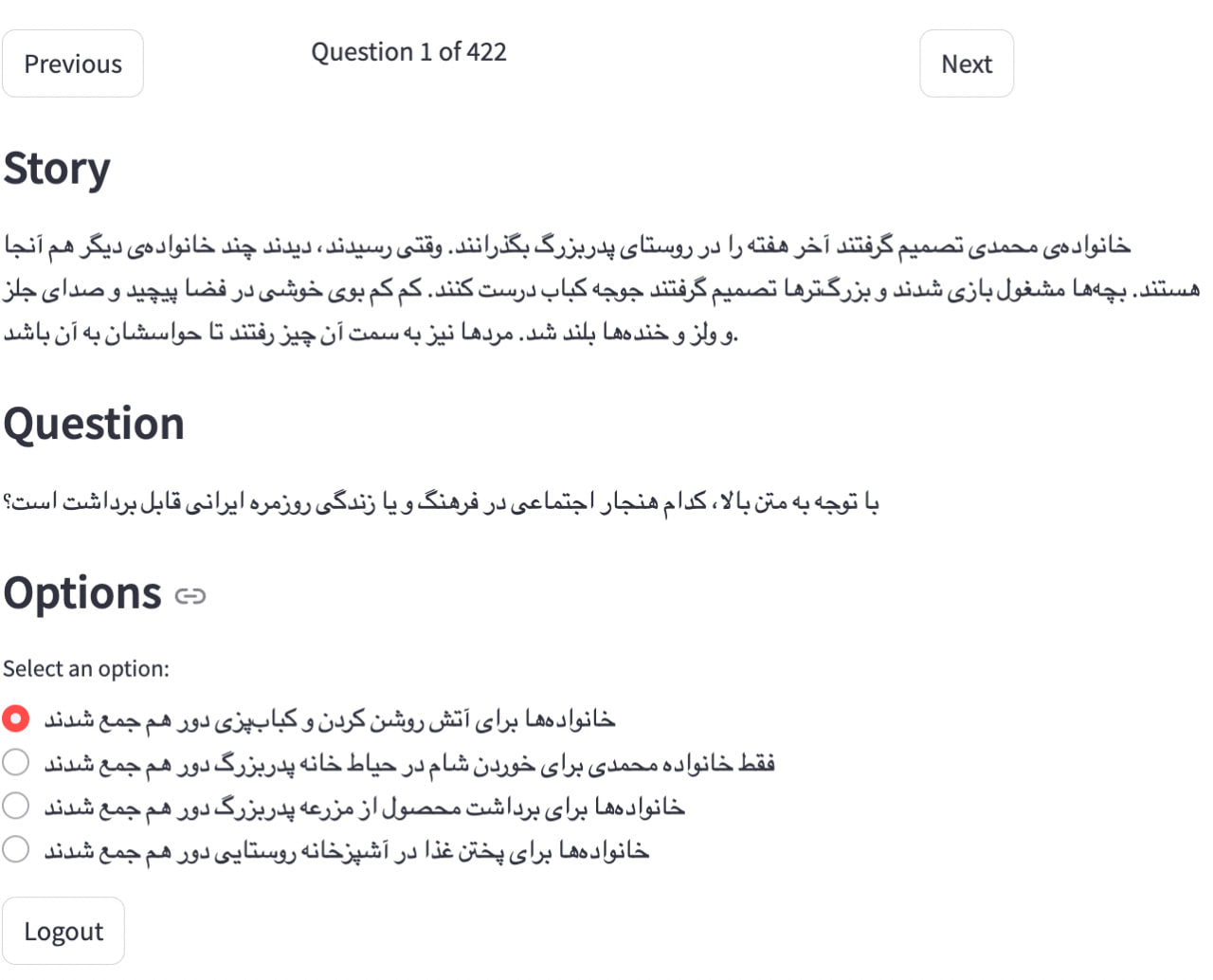}
    \caption{User interface for human baseline benchmarking on the dataset. Participants are presented with a story implying a cultural concept, followed by a comprehension question about the concept and multiple-choice options. The task requires users to select the most appropriate answer based on the annotation guidelines provided.}
    \label{fig:ui_human}
\end{figure}

\begin{figure}[!h]
    \centering
    \includegraphics[ width=1\linewidth]{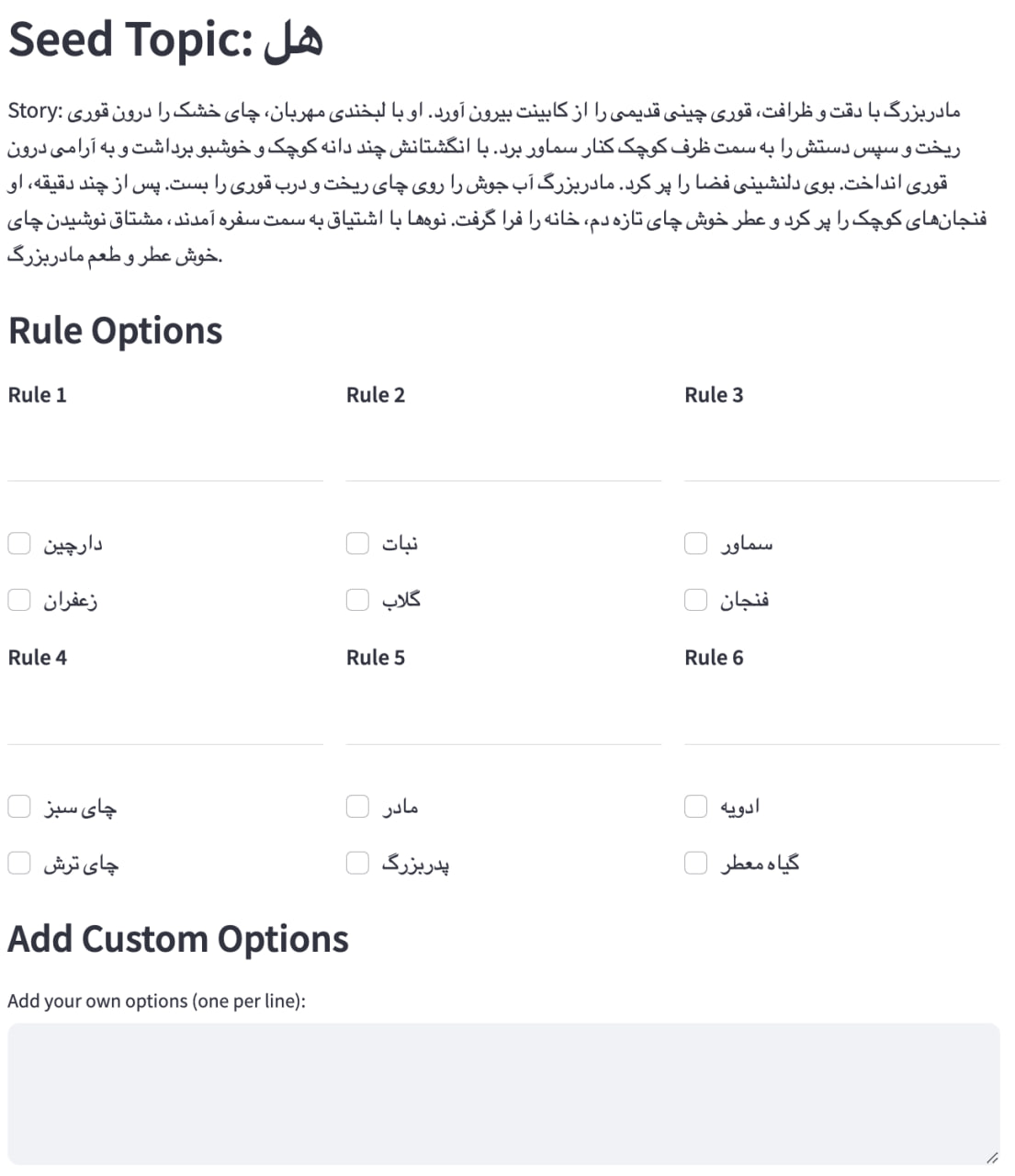}
    \caption{User interface for distractor selection. Participants are presented with 12 choices (2 generated per heuristic rule) and are tasked with selecting 3 distractors. Users can either choose from the provided options or add a new distractor manually if necessary.}
    \label{fig:ui_choice}
\end{figure}

\begin{figure}[!h]
    \centering
    \includegraphics[width=1\linewidth]{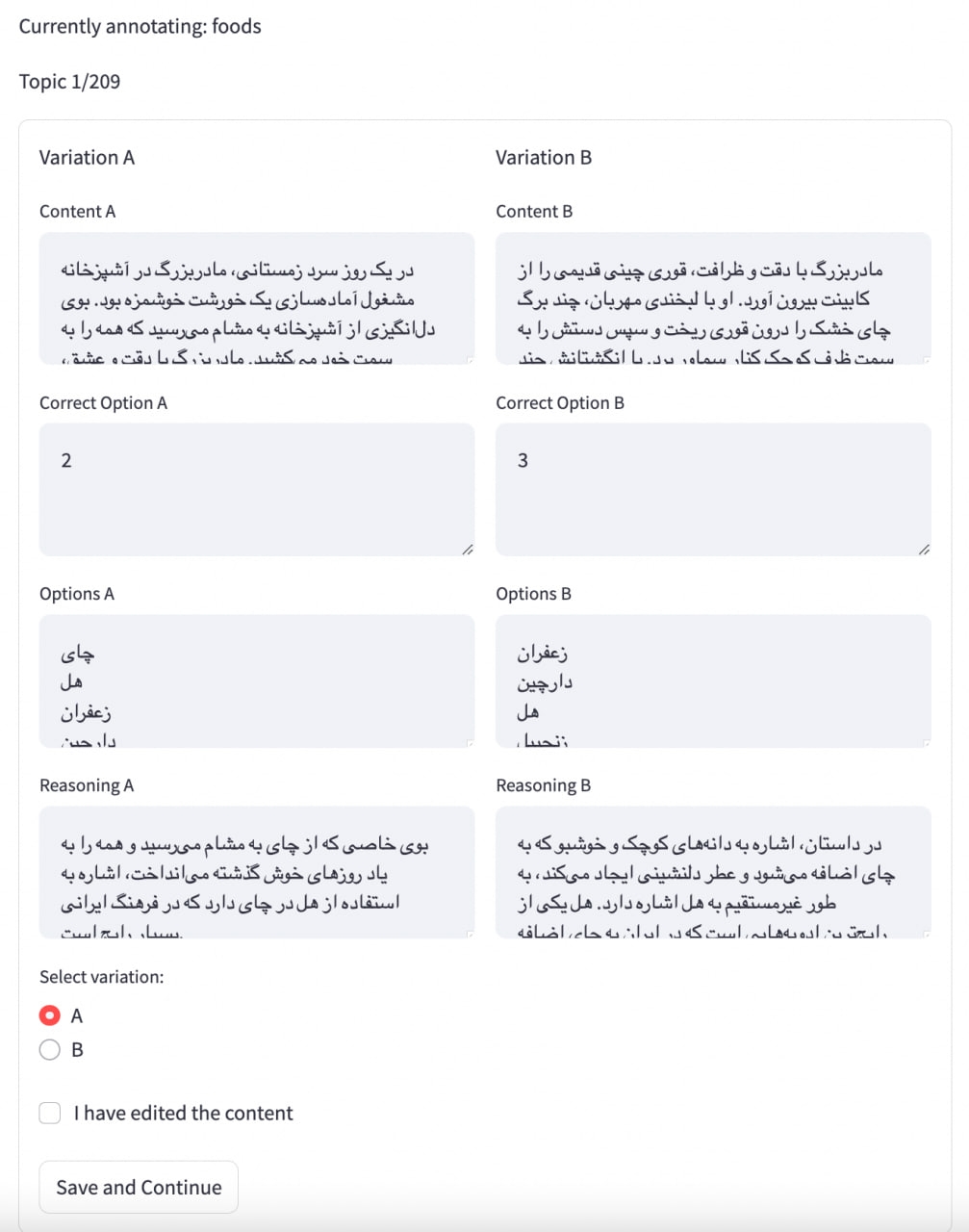}
    \caption{User interface for story selection and refinement task. Users are presented with two story variants generated from the same seed topic by different language models (Sonnet-3.5 and GPT-4o), along with default options and correct responses. Users can edit the content, select their preferred variant, and indicate if they made any modifications to the generated text. The source model for each variant is not disclosed to users during the task.}
    \label{fig:ui_story}
\end{figure}

We have developed various user interfaces for various steps of our work. Some of them are shown in Figures~\ref{fig:ui_human},~\ref{fig:ui_choice} and~\ref{fig:ui_story}.

\label{appendix:uis}

\section{Full Benchmark Results}
\label{appendix:full_bench}
\begin{figure}[!htp]
    \centering
    \includegraphics[width=1\linewidth]{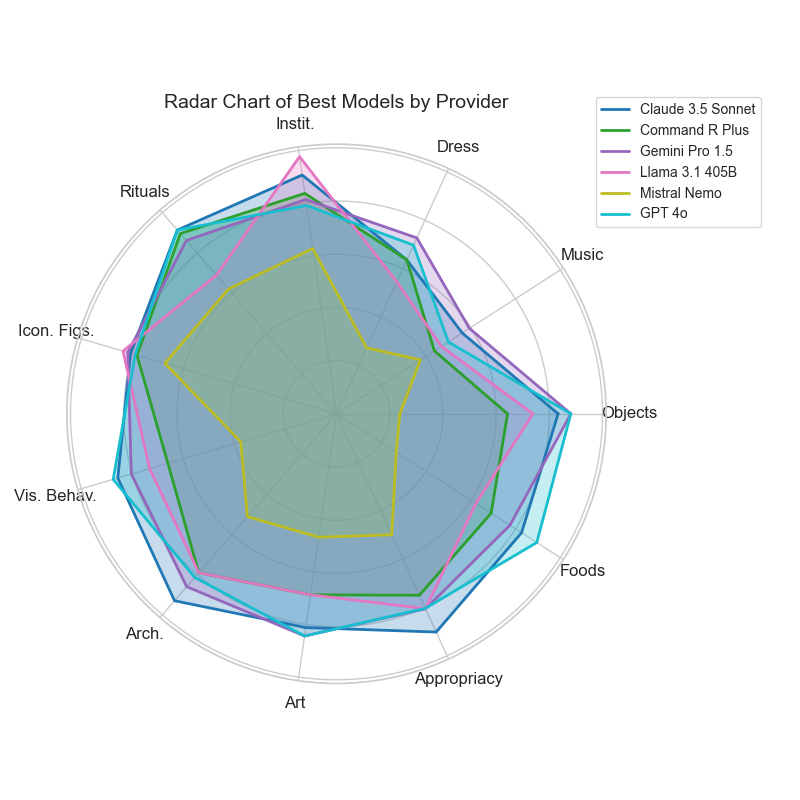}
    \caption{Radar Chart for best performing models of each family across categories.}
    \label{fig:radar_chart_flagship}
\end{figure}

\begin{figure*}[t!]
    \centering
    \includegraphics[trim={0 1cm 4cm 0},clip,width=0.8\textwidth]{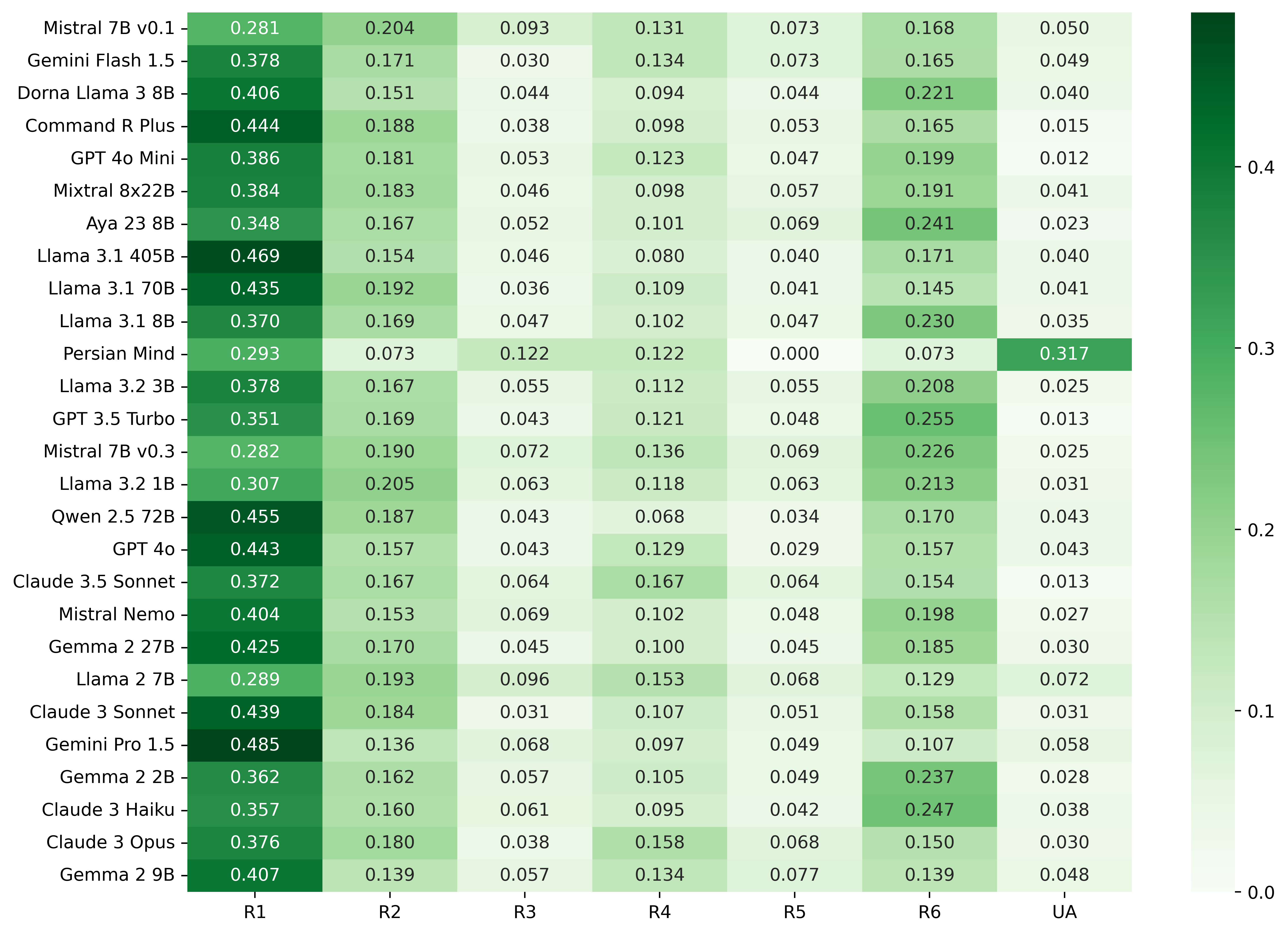}  % 
    \caption{Full heatmap on the effectiveness of distractor options created by different heuristic rules in misleading models. }
    \label{fig:distractor_full}
\end{figure*}

This appendix includes important visual and tabular data that complement our analysis. Figure~\ref{fig:radar_chart_flagship} provides a graphical representation of the performance metrics of the flagship member from each model family across various categories. Figure~\ref{fig:distractor_full} presents the full heatmap on the effectiveness of distractor options created by different heuristic rules in misleading models. Furthermore, Table~\ref{tab:combined_full_model_comparison} presents a comprehensive comparison of the accuracy of all models evaluated across all categories. These references highlight the detailed performance insights of our study.

\begin{sidewaystable*}[p]
\centering
\small
\resizebox{\textwidth}{!}{
\setlength{\tabcolsep}{10pt}
\begin{tabular}{llllllllllllll}
\toprule
Model & Objects & Music & Dress & Instit. & Rituals & Icon. Figs. & Vis. Behav. & Arch. & Art & Appropriacy & Foods & Macro Avg. & \# Params \\
\midrule
Claude 3.5 Sonnet & 0.833 & 0.562 & 0.636 & 0.907 & 0.914 & 0.807 & 0.857 & 0.930 & 0.812 & 0.902 & 0.827 & 0.817 & N/A \\
GPT 4o & 0.881 & 0.500 & 0.697 & 0.791 & 0.914 & 0.789 & 0.875 & 0.814 & 0.844 & 0.804 & 0.895 & 0.800 & N/A \\
Gemini Pro 1.5 & 0.881 & 0.594 & 0.727 & 0.814 & 0.862 & 0.818 & 0.804 & 0.860 & 0.844 & 0.806 & 0.775 & 0.799 & N/A \\
Claude 3 Opus & 0.810 & 0.375 & 0.727 & 0.953 & 0.966 & 0.836 & 0.839 & 0.814 & 0.844 & 0.861 & 0.702 & 0.793 & N/A \\
Command R Plus & 0.643 & 0.438 & 0.636 & 0.837 & 0.897 & 0.782 & 0.661 & 0.791 & 0.688 & 0.750 & 0.691 & 0.710 & 104B \\
Gemini Flash 1.5 & 0.786 & 0.500 & 0.667 & 0.953 & 0.724 & 0.800 & 0.750 & 0.698 & 0.719 & 0.778 & 0.670 & 0.731 & N/A \\
$\dagger$\textit{GPT 4o} & 0.643 & 0.500 & 0.667 & 0.791 & 0.793 & 0.727 & 0.696 & 0.558 & 0.719 & 0.750 & 0.759 & 0.691 & N/A \\
Llama 3.1 405B & 0.738 & 0.469 & 0.545 & 0.977 & 0.690 & 0.836 & 0.732 & 0.791 & 0.688 & 0.806 & 0.623 & 0.718 & 405B \\
$\dagger$\textit{Claude 3.5 Sonnet} & 0.548 & 0.406 & 0.636 & 0.791 & 0.793 & 0.745 & 0.643 & 0.581 & 0.750 & 0.806 & 0.707 & 0.673 & N/A \\
Claude 3 Sonnet & 0.595 & 0.531 & 0.606 & 0.907 & 0.759 & 0.691 & 0.750 & 0.674 & 0.656 & 0.750 & 0.565 & 0.680 & N/A \\
GPT 4o Mini & 0.738 & 0.469 & 0.545 & 0.674 & 0.621 & 0.691 & 0.589 & 0.674 & 0.656 & 0.722 & 0.686 & 0.642 & N/A \\
$\dagger$\textit{Gemini Pro 1.5} & 0.619 & 0.438 & 0.545 & 0.791 & 0.759 & 0.727 & 0.696 & 0.581 & 0.625 & 0.778 & 0.639 & 0.654 & N/A \\
$\dagger$\textit{Llama 3.1 405B} & 0.571 & 0.375 & 0.545 & 0.791 & 0.828 & 0.691 & 0.661 & 0.581 & 0.656 & 0.806 & 0.654 & 0.651 & 405B \\
Llama 3.1 70B & 0.690 & 0.344 & 0.545 & 0.930 & 0.759 & 0.673 & 0.643 & 0.814 & 0.656 & 0.750 & 0.602 & 0.673 & 70B \\
Gemma 2 9B & 0.643 & 0.438 & 0.727 & 0.814 & 0.586 & 0.727 & 0.661 & 0.744 & 0.781 & 0.778 & 0.524 & 0.675 & 9B \\
Gemma 2 27B & 0.714 & 0.406 & 0.545 & 0.837 & 0.655 & 0.727 & 0.732 & 0.767 & 0.594 & 0.778 & 0.592 & 0.668 & 27B \\
Qwen 2.5 72B & 0.571 & 0.406 & 0.606 & 0.767 & 0.655 & 0.709 & 0.607 & 0.651 & 0.656 & 0.639 & 0.539 & 0.619 & 72B \\
$\dagger$\textit{Command R Plus} & 0.524 & 0.312 & 0.333 & 0.674 & 0.724 & 0.709 & 0.661 & 0.581 & 0.562 & 0.694 & 0.607 & 0.580 & 104B \\
Claude 3 Haiku & 0.548 & 0.438 & 0.424 & 0.744 & 0.621 & 0.691 & 0.554 & 0.674 & 0.656 & 0.667 & 0.440 & 0.587 & N/A \\
Mistral Nemo & 0.238 & 0.375 & 0.273 & 0.628 & 0.621 & 0.673 & 0.375 & 0.512 & 0.469 & 0.500 & 0.267 & 0.448 & 12B \\
Llama 3.1 8B & 0.286 & 0.250 & 0.364 & 0.791 & 0.345 & 0.527 & 0.446 & 0.465 & 0.500 & 0.611 & 0.304 & 0.444 & 8B \\
Dorna Llama 3 8B & 0.357 & 0.250 & 0.394 & 0.721 & 0.379 & 0.509 & 0.446 & 0.442 & 0.469 & 0.556 & 0.314 & 0.440 & 8B \\
Aya 23 8B & 0.381 & 0.312 & 0.273 & 0.721 & 0.345 & 0.509 & 0.321 & 0.512 & 0.344 & 0.528 & 0.251 & 0.409 & 8B \\
Mixtral 8x22B & 0.405 & 0.188 & 0.333 & 0.535 & 0.241 & 0.473 & 0.339 & 0.465 & 0.406 & 0.583 & 0.298 & 0.388 & 8x22B \\
Gemma 2 2B & 0.190 & 0.281 & 0.303 & 0.465 & 0.414 & 0.345 & 0.339 & 0.558 & 0.344 & 0.361 & 0.225 & 0.348 & 2B \\
GPT 3.5 Turbo & 0.333 & 0.188 & 0.303 & 0.488 & 0.310 & 0.364 & 0.250 & 0.256 & 0.344 & 0.417 & 0.293 & 0.322 & N/A \\
Llama 3.2 3B & 0.214 & 0.219 & 0.121 & 0.465 & 0.345 & 0.273 & 0.161 & 0.233 & 0.312 & 0.333 & 0.194 & 0.261 & 3B \\
Mistral 7B v0.3 & 0.095 & 0.062 & 0.121 & 0.256 & 0.069 & 0.182 & 0.125 & 0.140 & 0.219 & 0.194 & 0.173 & 0.149 & 7B \\
Mistral 7B v0.1 & 0.071 & 0.062 & 0.182 & 0.326 & 0.310 & 0.127 & 0.196 & 0.116 & 0.094 & 0.083 & 0.126 & 0.154 & 7B \\
Llama 2 7B & 0.048 & 0.031 & 0.242 & 0.302 & 0.207 & 0.127 & 0.125 & 0.116 & 0.062 & 0.083 & 0.084 & 0.130 & 7B \\
Llama 3.2 1B & 0.024 & 0.000 & 0.061 & 0.070 & 0.172 & 0.018 & 0.018 & 0.163 & 0.125 & 0.028 & 0.021 & 0.064 & 1B \\
PersianMind v1.0 & 0.024 & 0.000 & 0.000 & 0.093 & 0.000 & 0.018 & 0.036 & 0.093 & 0.094 & 0.000 & 0.000 & 0.033 & 7B \\
\bottomrule
\end{tabular}
}
\caption{Comprehensive performance comparison of various models based on their accuracy over \textsc{PerCul}. Models with $\dagger$ next to their name are evaluated on English translation of \textsc{PerCul}.}
\label{tab:combined_full_model_comparison}
\end{sidewaystable*}

\section{Distractor Analysis Examples}
\label{appendix:distractor_examples}
Here we go over some examples of distractors, corresponding stories and the distracted explanations of models.

\begin{table*}[ht]
\centering
\small
\begin{tabular}{p{0.35\textwidth}p{0.15\textwidth}p{0.35\textwidth}}
\hline
\textbf{Story} & \textbf{Option (Rule)} & \textbf{Model Distraction Description} \\
\hline
\multirow{4}{0.35\textwidth}{On a warm summer night, the people of a small village by the sea had gathered together. The sound of drums and various instruments was in the air, and everyone was dancing together. An old man with a white beard and eyes full of memories was sitting in a corner, smiling at the young people who were enthusiastically responding to the sound of music. Children were running happily among the crowd, and women were dancing beautifully in their colorful dresses. From time to time, the sound of a flute could be heard, giving the crowd a special atmosphere. These celebrations were always an excuse to get together and have fun, and no one wanted these beautiful moments to end.} 
& \vspace{2mm}Rural wedding celebration (R2)\vspace{2mm} & \vspace{2mm}This rural wedding celebration was an excuse to be together and celebrate, and no one wanted these beautiful moments to end.\vspace{2mm} \\
\cline{2-3}
& \vspace{2mm}Traditional coastal dance (R1)\vspace{2mm} & \vspace{2mm}This is a description of a traditional coastal dance and indicates that the associated choice number is 2.\vspace{2mm} \\
\cline{2-3}
& \vspace{2mm}Traditional Iranian music (R3)\vspace{2mm} & \vspace{2mm}According to the description of the text, the elements of traditional Iranian music and culture that can be understood are traditional Iranian music. The text refers to the sound of drums, various instruments, and the neyban, all of which are considered part of traditional Iranian music.\vspace{2mm} \\
\cline{2-3}
& \vspace{2mm}Bandari music (Correct)\vspace{2mm} & \vspace{2mm}-\vspace{2mm} \\
\hline
\multirow{4}{0.35\textwidth}{With a kind smile, the grandmother put a large pot on the stove and started cooking. A pleasant smell filled the house. She patiently stirred the contents of the pot. The soft, cooked grains that were the main feature of this soup were mixed with fragrant vegetables. The grandchildren eagerly waited for the hot and nutritious food to be prepared. The grandmother brought the clay bowls and carefully poured the contents of the pot into them. Then she added some delicious white liquid along with garlic to each bowl, which doubled the taste of the food. Everyone gathered around the table to enjoy this traditional and special dish.}
& \vspace{2mm}Sirdagh (R3)\vspace{2mm} & \vspace{2mm}Sirdagh is relevant to Iranian culture and daily life.\vspace{2mm} \\
\cline{2-3}
& \vspace{2mm}Abgousht (R2)\vspace{2mm} & \vspace{2mm}No model was distracted with this option\vspace{2mm} \\
\cline{2-3}
& \vspace{2mm}Doogh Ash (R4)\vspace{2mm} & \vspace{2mm}Based on the textual descriptions, it seems that the dish that the grandmother is cooking is a type of soup. The reference to soft, cooked grains, aromatic vegetables, and the addition of a white liquid (probably doogh or curd) and garlic at the end all indicate doogh soup. Doogh soup is one of the traditional and popular types of soup in Iranian food culture.\vspace{2mm} \\
\cline{2-3}
& \vspace{2mm}Barley Ash (Correct)\vspace{2mm} & \vspace{2mm}-\vspace{2mm} \\
\hline
\multirow{4}{0.35\textwidth}{Isfahani grandmother was meticulously preparing a special meal for the family. She first cooked the meat with onions and a little turmeric, then pounded it well until it was soft and tender. In another pot, she put a mixture of dairy and eggs on low heat, stirring constantly until it reached a consistency. When the mixture came to a boil, she added the pounded meat to it and stirred patiently until the ingredients were thoroughly mixed and elastic. Finally, she added a little saffron and rose water to give the food a pleasant aroma. She put the food in the refrigerator to cool and be ready for the evening. The grandchildren were eagerly waiting to enjoy this special and cool meal with almond slices on top.}
& \vspace{2mm}Eggplant Curd (R3)\vspace{2mm} & \vspace{2mm}No model was distracted with this option\vspace{2mm} \\
\cline{2-3}
& \vspace{2mm}Isfahani Khoresht Gheymeh (R5)\vspace{2mm} & \vspace{2mm}Given the description of preparing the food with ingredients such as meat, onion, turmeric, dairy, eggs, saffron, and rose water, this dish appears to be Isfahani Gheemeh Stew, which is one of the traditional and popular dishes in Iranian culture and daily life.\vspace{2mm} \\
\cline{2-3}
& \vspace{2mm}Fesenjan (R4)\vspace{2mm} & \vspace{2mm}No model was distracted using this option\vspace{2mm} \\
\cline{2-3}
& \vspace{2mm}Khoresht Mast (Correct)\vspace{2mm} & \vspace{2mm}-\vspace{2mm} \\
\hline
\end{tabular}
\caption{Examples of stories used in cultural evaluation with their corresponding options and model distraction descriptions.}
\label{tab:story_examples}
\end{table*}

\end{document}